\documentclass{article}

\PassOptionsToPackage{numbers, compress}{natbib}
\usepackage[final]{neurips_2022}

\usepackage[utf8]{inputenc} % allow utf-8 input
\usepackage[T1]{fontenc}    % use 8-bit T1 fonts
\usepackage{hyperref}       % hyperlinks
\usepackage{url}            % simple URL typesetting
\usepackage{booktabs}       % professional-quality tables
\usepackage{amsfonts}       % blackboard math symbols
\usepackage{nicefrac}       % compact symbols for 1/2, etc.
\usepackage{microtype}      % microtypography
\usepackage{xcolor}         % colors

\usepackage[export]{adjustbox}

\usepackage{bm}
\usepackage{placeins}
\usepackage{graphicx}
\usepackage{amsmath}
\usepackage{amssymb}
\usepackage{amsthm}
\usepackage{caption}
\usepackage{subcaption}

\DeclareMathOperator{\E}{\mathbb{E}}

\newcommand{\card}[1]{\lvert#1\rvert}

\newtheorem{propo}{Proposition}
\theoremstyle{definition}
\newtheorem{definition}{Definition}
\newtheorem{remark}{Remark}

\newenvironment{nospaceflalign*}
{
	\setlength{\abovedisplayskip}{0pt}\setlength{\belowdisplayskip}{0pt}%
	\csname flalign*\endcsname}
{\csname endflalign*\endcsname\ignorespacesafterend}

\newcommand{\Pb}{\mathbb{P}}
\newcommand{\bC}{\mathbf{C}}
\newcommand{\bP}{\mathbf{P}}
\newcommand{\bh}{\mathbf{h}}
\newcommand{\bff}{\mathbf{f}}
\newcommand{\bg}{\mathbf{g}}
\newcommand{\bG}{\mathbf{G}}
\newcommand{\bzero}{\mathbf{0}}
\newcommand{\blambda}{\bm{\lambda}}
\newcommand{\bmeta}{\bm{\eta}}
\newcommand{\bkappa}{\bm{\kappa}}
\newcommand{\bmu}{\bm{\mu}}

\newcommand{\bgamma}{\bm{\gamma}}
\newcommand{\bphi}{\bm{\phi}}
\newcommand{\bpsi}{\bm{\psi}}
\newcommand{\OT}{\text{\normalfont{OT}}}
\newcommand{\OTF}{\text{\normalfont{OTF}}}
\newcommand{\OTe}{\OT_{\epsilon}}
\newcommand{\OTFe}{\OTF_{\epsilon}}
\newcommand{\tOTFe}{\OTF_{\epsilon}^{0}}
\newcommand{\OTFRe}{\text{\normalfont{OTFR}}_{\epsilon}}

\newcommand{\Pif}{\Pi^\mathcal{F}}
\newcommand{\dF}{d_{\mathcal{F}}}
\newcommand{\dS}{d_{\mathcal{S}}}
\newcommand{\dX}{d_{\mathcal{X}}}
\newcommand{\distrimage}[1]{\includegraphics[width=.175\linewidth,valign=m]{./figures/distributions/#1}}

\DeclareMathOperator*{\argmax}{argmax}
\DeclareMathOperator*{\argmin}{argmin}

\title{Optimal Transport of Classifiers to Fairness}

\author{Maarten Buyl\\
	Ghent University\\
	\texttt{maarten.buyl@ugent.be}
	\And Tijl De Bie \\
	Ghent University\\
	\texttt{tijl.debie@ugent.be}}

\begin{document}

\maketitle

%
%
%%%%%%%%%%%%%%%%%%%%%%%%%%%%%%%%%%%%%%%%%%%%%%%%%%%%%%%%%%%%
\begin{abstract}
In past work on fairness in machine learning, the focus has been on forcing the prediction of classifiers to have similar statistical properties for people of different demographics. To reduce the violation of these properties, fairness methods usually simply rescale the classifier scores, ignoring similarities and dissimilarities between members of different groups. Yet, we hypothesize that such information is relevant in quantifying the unfairness of a given classifier. To validate this hypothesis, we introduce Optimal Transport to Fairness (OTF), a method that quantifies the violation of fairness constraints as the smallest Optimal Transport cost between a probabilistic classifier and any score function that satisfies these constraints. For a flexible class of linear fairness constraints, we construct a practical way to compute OTF as a differentiable fairness regularizer that can be added to any standard classification setting. Experiments show that OTF can be used to achieve an improved trade-off between predictive power and fairness.
\end{abstract}
%%%%%%%%%%%%%%%%%%%%%%%%%%%%%%%%%%%%%%%%%%%%%%%%%%%%%%%%%%%

%%%%%%%%%%%%%%%%%%%%%%%%%%%%%%%%%%%%%%%%%%%%%%%%%%%%%%%%%%%
\section{Introduction}
%%%%%%%%%%%%%%%%%%%%%%%%%%%%%%%%%%%%%%%%%%%%%%%%%%%%%%%%%%%%

Machine learning methods are increasingly being deployed for automated decision making due to the potential benefits in terms of efficiency and effectiveness. Yet, making automated decisions about \textit{people} comes with substantial legal and ethical risks, as evidenced by more and more cases of undesirable algorithmic discrimination \cite{larsonHowWeAnalyzed2016, dastinAmazonScrapsSecret2018, chowdhurySharingLearningsOur2021} with respect to the protected traits of individuals. Even decisions that appear neutral because they do not use such sensitive information, may still disproportionately affect certain groups indirectly \cite{wachterBiasPreservationMachine2020}.

The research field of \textit{fairness} in machine learning \cite{mehrabininarehSurveyBiasFairness2021} therefore studies ways in which a model's discrimination with respect to a person's \textit{sensitive features} can be reduced or removed. Many notions of fairness have been proposed for binary classification tasks \cite{vermaFairnessDefinitionsExplained2018b}. A popular example is \textit{demographic parity}, which is expressed as a constraint that enforces equality between the rates at which positive decisions are made for each protected group \cite{caldersBuildingClassifiersIndependency2009}.

\paragraph{Motivation}
For a given fairness notion, we would like to tune a model that satisfies the constraint that expresses the notion. To this end, it can be practical to quantify the violation of this constraint as a fairness regularization term that can be added to any probabilistic classifier \cite{kamishimaFairnessawareLearningRegularization2011}. Such an approach provides an incentive to the model to learn parameters that result in fair probability scores, e.g. by learning to ignore highly biased features. A straightforward fairness regularizer is the norm of the difference between the quantities that should be equal according to the fairness notion \cite{zafarFairnessConstraintsFlexible2019, padalaFNNCAchievingFairness2021}. 

However, such a quantification of unfairness may be limiting, as it only considers the statistical properties of the model's output, while ignoring the input features for which those scores were computed. We argue that, to quantify the unfairness of a model as a fairness regularizer, it can be beneficial to more strongly consider data points with similar features but significant unfair discrepancies in probability scores. To this end, we employ \textit{Optimal Transport} (OT) \cite{peyreComputationalOptimalTransport2020} as a way to measure these discrepancies while taking feature similarity into account.

%\newpage
\paragraph{Contributions}
\begin{itemize}
	\item We propose to quantify the unfairness of a probabilistic classifier as the \textit{Optimal Transport to Fairness} (OTF) cost, which is defined as the smallest OT cost between the score function of the classifier and any fair score function over the same data.
	\item We make novel derivations to compute OTF as a \textit{differentiable fairness regularizer} that can be added to the training loss of any probabilistic classifier and is efficiently computed for popular notions of fairness such as \textit{demographic parity} and \textit{equalised odds}. Our approach is capable of handling multiple sensitive variables, that can be categorical or continuous.
	\item In experiments, our method shows its benefit of increased flexibility over other OT methods. In some cases, it also achieves a trade-off between predictive power and fairness that is significantly more effective for equalized odds.
\end{itemize}

%\paragraph{Outline}
%In Sec.~\ref{sec:background} we recall the standard context of fair binary classification, discuss the practical use of linear fairness notions and introduce the OT problem. With these preliminaries, we gradually construct the OTF method in Sec.~\ref{sec:otf}. We compare with related work in Sec.~\ref{sec:related} and end with experiment results in Sec.~\ref{sec:experiments}.

%
%
%%%%%%%%%%%%%%%%%%%%%%%%%%%%%%%%%%%%%%%%%%%%%%%%%%%%%%%%%%%%
\section{Background}\label{sec:background}
%%%%%%%%%%%%%%%%%%%%%%%%%%%%%%%%%%%%%%%%%%%%%%%%%%%%%%%%%%%%
\subsection{Fair Classification}\label{sec:fair_class}
Let $\mathcal{Z} \triangleq \mathcal{X} \times \mathcal{S} \times \{0, 1\}$ denote the sample space, from which we draw samples $Z \triangleq (X, S, Y)$ and try to estimate the \textit{output} label $Y \in \{0, 1\}$ from \textit{input} features $X \in \mathcal{X}$, without discriminating with respect to \textit{sensitive} features $S \in \mathcal{S}$. It is assumed that $\mathcal{X} \subset \mathbb{R}^{\dX}$ and $\mathcal{S} \subset \mathbb{R}^{\dS}$.% where $\dX$ and $\dS$ are the dimensionality of $\mathcal{X}$ and $\mathcal{S}$.
We sample binary predictions $\hat{Y} \in \{0, 1\}$ from a probabilistic classifier $h: \mathcal{X} \to [0, 1]$ that assigns a score $h(X)$ to the belief that a sample with features $X$ belongs to the positive class. %The Bernoulli distribution of $\hat{Y}$ is thus fully specified by $h(X)$. 
Our general goal in fair classification is to minimize a loss function $\mathcal{L}_Y(h)$, in our case the cross-entropy between $h(X)$ and $Y$, while also scoring well on a fairness measurement of $h(X)$.
%We aim to perform binary predictions $\hat{Y} \in \{0, 1\}$ that are similar to the true value of $Y$. Let $h: \mathcal{X} \to [0, 1]$ denote a probabilistic classifier that assigns a score $h(X)$ to the belief that a sample with features $X$ belongs to the positive class. Thus, the distribution of the prediction $\hat{Y} \mid X$ is fully specified as the Bernoulli distribution with parameter $h(X)$.
%In what follows, we will interchangeably use the notation $h$ to refer to a probabilistic classifier, its score function or the measure tied to that score function (discussed in Sec. \ref{sec:tying_score}). 

%\subsection{Fairness Constraints}\label{sec:constraints}

\subsubsection{Group Fairness}
The fairness of classifiers is commonly enforced through some notion of independence from sensitive information $S \in \mathcal{S}$. In particular, \textit{group fairness} is concerned with constraints over statistical measurements of this independence over all members of a protected group \cite{mehrabininarehSurveyBiasFairness2021}. For now, assume that $\mathcal{S}$ consists of a single, categorical feature with $\dS$ values. We denote $S \in \mathcal{S}$ as the one-hot encoding\footnote{For example, if $\mathcal{S}$ only contains two categories, then we denote $\mathcal{S} = \{[1, 0], [0, 1]\}$ with $\card{\mathcal{S}} = \dS = 2$.} with values $S_k \in \{0, 1\}$. This encoding is indexed by values from $[\dS] = \{0, ..., \dS - 1\}$.

Assuming that the predictions $\hat{Y} \mid X$ are randomly sampled from a probabilistic classifier $h(X)$, then the traditional fairness notion of \textit{demographic parity} (DP) is equivalent to enforcing zero covariance between $h(X)$ and each group in $S$, which we refer to as Probabilistic Demographic Parity (PDP):
\begin{flalign}\label{eq:cov}
	\text{\qquad (PDP)}&&
	\forall k \in [\dS]:
	\E_Z\left[h(X)S_k\right] - \E_Z[h(X)]\E_Z[S_k] = 0.&&
	\phantom{\text{\qquad (PDP)}}
\end{flalign}

\begin{remark}
If $\hat{Y}$ is not sampled from $h(X)$ but instead decided by a threshold (e.g. $\hat{Y} = 1$ if $h(X) > 0.5$), then Eq. (\ref{eq:cov}) is a relaxation of the actual DP notion and may therefore admit unfair models \cite{lohausTooRelaxedBe2020}. However, such a threshold introduces a discontinuity with respect to the input, making it difficult to efficiently find fair classifiers in practice \cite{taskesenStatisticalTestProbabilistic2020, agarwalReductionsApproachFair2018}. Moreover, by considering the probabilities directly, we account for the uncertainty expressed by the probabilistic classifier.
\end{remark}

The notation in Eq. (\ref{eq:cov}) is easily extended to fairness notions that condition on other variables. For example, the \textit{equalized odds} (EO) \cite{hardtEqualityOpportunitySupervised2016} fairness notion conditions DP on the actual label $Y$. We apply EO to probabilistic classifiers by conditioning the covariance constraint of PDP on the one-hot encoding of $Y$ and retrieve the following expression for Probabilistic Equalized Odds (PEO):
\begin{flalign}\label{eq:cond_cov}
	\text{\qquad (PEO)}&&
	\forall l \in \{0, 1\}, \forall k \in [\dS]: \E_Z\left[h(X)S_kY_l\right] - \E_Z[h(X)Y_l]\E_Z[S_kY_l] = 0.&&
	\phantom{\text{\qquad (PDP)}}
\end{flalign}

\subsubsection{Linear Fairness}\label{sec:lin_fair}
The PDP and PEO covariance constraints in Eq. (\ref{eq:cov}) and Eq. (\ref{eq:cond_cov}) are practical, because they can be written as linear sums over the probabilities given by the classifier $h$. As was done in past work \cite{agarwalReductionsApproachFair2018}, we are generally interested in all fairness notions that can be enforced through such linear constraints.

\begin{definition}[Linear fairness notion]\label{def:lin_fairness}
	A notion of fairness is a \textit{linear fairness notion} when the set $\mathcal{F}$ of all score functions $f: \mathcal{X} \to [0, 1]$ that satisfy it is given by
	\begin{equation}\label{eq:lin_fairness}
		\mathcal{F} \triangleq \{f: \mathcal{X} \to [0, 1]: \E_Z\left[\bg(Z)f(X) \right] = \bzero_{\dF}\}
	\end{equation}
	with $\bzero_{\dF}$ a vector of $\dF$ zeros and the vector-valued function $\bg(Z) : \mathcal{Z} \to \mathbb{R}^{\dF}$ constant w.r.t. $f(X)$.
\end{definition}

\begin{table}[]
	\centering
	\caption{Mapping of binary fairness notions onto the $\bg$ functions of their probabilistic version to be written as linear fairness notions.}
	\label{tab:lin_fair}
	\begin{tabular}{l|c|c}
		& Predictions Constraint & Linear Fairness Notion \\ \hline
		(P)DP & $\Pb(\hat{Y} = 1 \mid S) = \Pb(\hat{Y} = 1)$ & $\bg_{k} = \frac{S_k}{\E_Z[S_k]} - 1$ \\ \hline
		(P)EO & $\Pb(\hat{Y} = 1 \mid S, Y) = \Pb(\hat{Y} = 1 \mid Y)$ & $\bg_{k+l\dS} = Y_l\left(\frac{S_k}{\E_Z[S_kY_l]} - 1\right)$ 
	\end{tabular}
\end{table}

PDP and PEO are clearly linear fairness notions, with their $\bg$ function listed in Table \ref{tab:lin_fair}. Linear constraints are practical because the set of fair score functions $\mathcal{F}$ is convex. Moreover, multiple linear fairness notions can be combined into a new linear fairness notion by concatenating their constraint vectors. Sensitive information that is inherently continuous, like age, can be considered as a single dimension in $S$, thereby allowing for a mix of categorical and continuous sensitive attributes.

%Some sensitive information, like age, is inherently continuous. In our setup, we therefore also allow for a mix of multiple categorical and continuous sensitive attributes by considering them as different dimensions in $S$. Categorical attributes then have as many dimensions as there are categories, while continuous attributes each consist of one dimension in $S$. 

However, under linear fairness, a non-linear dependence between a score function $f(X)$ and \textit{continuous} sensitive attributes $S$ can remain undetected. Also, some well-known fairness notions are not linear. For example, notions related to \textit{sufficiency} involve conditioning on $f(X)$, meaning that $f(X)$ also shows up in the $\bg$ function. To this end, approaches have been proposed \cite{celisClassificationFairnessConstraints2019} that can approximate classifiers with non-linear fairness through a reduction to problems with linear constraints. 

\subsection{Optimal Transport of Score Functions}\label{sec:ot}
Optimal Transport (OT) \cite{peyreComputationalOptimalTransport2019} theory considers the problem of moving a mass from one measure to another at the smallest possible total cost. Here, we let every score function $f$ correspond with the measure $\theta_f$ defined over the \textit{input} space $\mathcal{X}$ endowed with the Borel $\sigma$-algebra: $\theta_f \triangleq \sum_{x \in \mathcal{D_X}} f(x)\delta_x$, with $\delta_x$ the Dirac measure and $\mathcal{D_X}$ the input features of samples in a collection $\mathcal{D}$. Note that the input space measure $\theta_f$ is not normalized, though this is not necessary to apply OT theory. In what follows, we implicitly consider the score functions $h$ and $f$ as their corresponding input space measures $\theta_h$ and $\theta_f$ when used in the OT problem. See the Appendix~\ref{sec:tying} 
for further clarification.

For a collection $\mathcal{D}$ of $n$ samples (e.g. a full dataset or only a batch from it), let $\bh$ and $\bff$ denote the $n$-dimensional vectors of score function values for all data points, i.e. $\bh_i = h(x_i)$ and $\bff_i = f(x_i)$. Furthermore, for a non-negative transport cost function $c$ defined over $\mathcal{X} \times \mathcal{X}$, let $\mathbf{C} \in \mathbb{R}_+^{n \times n}$ represent the matrix of cost terms, i.e. $\bC_{ij} = c(x_i, x_j)$. Similarly, with $\pi(x_i, x_j)$ the \textit{coupling} that reflects how much score mass was transported from $x_i$ to $x_j$, define the matrix $\bP \in \mathbb{R}_+^{n \times n}$ with $\bP_{ij} = \pi(x_i, x_j)$. The OT cost is then simplified to
\begin{equation}\label{eq:ot_vect}
	\OT(h, f) = \min_{\bP \in \Pi(h, f)} \langle \bC, \bP \rangle
\end{equation}
\begin{nospaceflalign*}
	\text{with} && \Pi(h, f) = \left\{\bP \in \mathbb{R}_+^{n \times n}: \bP \mathbf{1}_n = \bh, \bP^T \mathbf{1}_n = \bff \right\}.&&
\end{nospaceflalign*}

where $\mathbf{1}_n$ is the $n$-dimensional vector of ones and $\langle \bC, \bP \rangle = \sum_{ij} \bC_{ij}\bP_{ij}$. %Equation (\ref{eq:ot_vect}) is referred to as the \textit{Wasserstein} distance if the cost function $c$ is a metric.

%
%
%%%%%%%%%%%%%%%%%%%%%%%%%%%%%%%%%%%%%%%%%%%%%%%%%%%%%%%%%%%%
\section{Optimal Transport to Fairness}\label{sec:otf}
%%%%%%%%%%%%%%%%%%%%%%%%%%%%%%%%%%%%%%%%%%%%%%%%%%%%%%%%%%%%
%Fairness constraints as in Sec.~\ref{sec:fair_class} can be applied to a classifier's score function $h$ to test whether the scores satisfy our notion of fairness. If $h$ is shown to be unfair, however, there are many ways to quantify this unfairness. To motivate our proposed definition of unfairness, we set up the following thought experiment. 

Assume that a set $\mathcal{F}$ is available that denotes all score functions which satisfy the required notion of fairness. We would like to quantify the unfairness of a classifier's score function $h$ as the amount of 'work' minimally required to make it a member of this set by measuring how far $h$ is from its fair \textit{projection} onto $\mathcal{F}$, i.e. the $f \in \mathcal{F}$ that is \textit{closest} to $h$. We measure this closeness as the OT cost between $h$ and $f$, as we then assign a higher unfairness cost to $h$ if scores need to be transported between highly dissimilar individuals in order to reach a fair function $f \in \mathcal{F}$.

%Assume that a set $\mathcal{F}$ is available that denotes all score functions which satisfy the required notion of fairness. If a classifier's score function $h$ is unfair, then it ought to be replaced by a fair score function $f \in \mathcal{F}$. Arguably, the most suitable $f$ for this purpose is the \textit{projection} of $h$ onto $\mathcal{F}$, i.e. the $f \in \mathcal{F}$ that is \textit{closest} to $h$. However, replacing $h$ by its fair projection may come at a cost, e.g. because the latter may not be as accurate in predicting the data as $h$ was. The unfairness of $h$ can thus be quantified at any point by measuring how far it is from its fair projection at that point. 

Thus, we propose to quantify unfairness as the Optimal Transport to Fairness (OTF) cost, i.e. the cost of the OT-based projection of $h$ onto $\mathcal{F}$:
\begin{equation}\label{eq:otf_intro}
	\OTF(h) = \min_{f \in \mathcal{F}} \OT(h, f).
\end{equation}

In what follows, we expand on the OTF method by constructing it in three steps.
\begin{enumerate}
	\item In Sec.~\ref{sec:otlf}, we directly express the $\OTF(h)$ objective as a linear programming problem by making the assumption that $\mathcal{F}$ is defined by a linear fairness notion as in Def.~\ref{def:lin_fairness}.
	\item In Sec.~\ref{sec:entr_reg}, we add entropic smoothing to $\OTF(h)$, thereby making the resulting $\OTFe(h)$ cost a differentiable fairness regularizer that can be computed efficiently.
	\item In Sec.~\ref{sec:otf_adjusted}, we address the fact that due to this smoothing, $\OTFe(h)$ does not necessarily equal zero for a fair $h$. To this end, we subtract the adjustment term $\OTFRe(h)$, resulting in the adjusted $\tOTFe(h)$ cost.
\end{enumerate}

Finally in Sec.~\ref{sec:min_otf}, we show how $\tOTFe(h)$ can be minimized w.r.t. (the parameters of) $h$.
%After having fully constructed the adjusted, regularized OTF cost, we finally propose how this measure of unfairness can be minimized 

For all derivations and proofs, we refer to Appendix~\ref{sec:proofs}.

\subsection{Optimal Transport to Linear Fairness}\label{sec:otlf}
To compute the minimization in Eq.~(\ref{eq:otf_intro}), first note that in the $\OT(h, f)$ cost in Eq.~(\ref{eq:ot_vect}), $f$ only shows up in the constraint $\bP^T \mathbf{1}_n = \bff$ on the column marginals of coupling matrix $\bP$. It therefore suffices to weaken this constraint to $\bP^T \mathbf{1}_n \in \mathcal{F}$. Next, recall that a \textit{linear} fairness notion is enforced through a vector of constraints on the expectation of $\mathbf{g}(Z)f(X)$. For a collection $\mathcal{D}$ of $n$ samples, let $\bG_{cj} = \mathbf{g}_c(z_j)$, i.e. $\bG \in \mathbb{R}^{\dF \times n}$ is the constraints matrix with a row for every constraint and a column for every data point. It defines the linear fairness notion for $\mathcal{D}$.

\begin{definition}[$\OTF$]\label{def:otf_lin}
	For a linear fairness notion expressed through constraints matrix $\bG$ for $n$ samples and with non-negative cost matrix $\bC$, the \textit{Optimal Transport to Fairness} cost for score function $h: \mathcal{X} \to [0, 1]$ is computed as
	\begin{equation}\label{eq:otf_lin}
		\OTF(h) = \min_{\bP \in \Pif(h)} \langle \bC, \bP \rangle
	\end{equation}
	\begin{nospaceflalign*}
		\text{with} && \Pif(h) = \left\{\bP \in \mathbb{R}_+^{n \times n}: \bP \mathbf{1}_n = \bh, \mathbf{G} \bP^T \mathbf{1}_n = \bzero_{\dF} \right\}.&&	
	\end{nospaceflalign*}
\end{definition}

The optimal coupling computed for $\OTF(h)$ implicitly transports the scores of $\bh$ to the fair vector $\bff$ given the coupling's column marginals (i.e. $\bff = \bP^T \mathbf{1}_n$). Note that for any coupling matrix that satisfies the first constraint ($\bP \mathbf{1}_n = \bh$), we have that $\bh^T \mathbf{1}_n = \mathbf{1}_n^T \bP^T \mathbf{1}_n = \bff^T \mathbf{1}_n$, meaning that $\bh$ and this implicitly found vector $\bff$ sum to the same value. Scores of $h$ are therefore only \textit{transported} to a fair score function and no extra scores are created or destroyed.

For $\bP_{ij} \geq 0$, the optimization problem in Eq.~(\ref{eq:otf_lin}) can be solved through linear programming. However, its solution is not differentiable with respect to the score function $h$ \cite{cuturiSinkhornDistancesLightspeed2013, frognerLearningWassersteinLoss2015}. It is thus impractical as a fairness regularization term.
%However, its solution coupling is sparse,
%since it is optimal to meet the linear constraint by greedily transporting scores among data points $x_i$ and $x_j$ where $\bC_{ij}$ is lowest until the budget given by $\bh_i$ at index $i$ runs out. Moreover, the solution is 
%which means $\OTF(h)$ is not directly differentiable with respect to the score function $h$ \cite{frognerLearningWassersteinLoss2015}. It is thus impractical as a fairness regularization term.

\subsection{Entropic Smoothing}\label{sec:entr_reg}
A well-known trick to make OT costs differentiable is to apply \textit{entropic regularization} or \textit{smoothing} \cite{cuturiSinkhornDistancesLightspeed2013, peyreComputationalOptimalTransport2020}, which is done by maximizing the entropy of the coupling while optimizing the OT cost. We apply the same principle to make OTF differentiable.

\begin{definition}[$\OTFe$]\label{def:otfe}
	For a linear fairness notion expressed through constraints matrix $\bG$ for $n$ samples and with non-negative cost matrix $\bC$, the \textit{smooth OTF} cost with smoothing strength $\epsilon > 0$ for score function $h: \mathcal{X} \to (0, 1]$ is computed as
	\begin{equation}\label{eq:otfe}
		\OTFe(h) = \min_{\bP \in \Pif(h)} \langle \bC, \bP \rangle - \epsilon H(\bP)
	\end{equation}
with $H(\bP) = - \sum_{ij} \bP_{ij} \left(\log\left(\bP_{ij}\right) - 1\right)$ the entropy of coupling $\bP$ and $\epsilon > 0$ a hyperparameter that regulates the strength of entropic smoothing.
	%with
	%\begin{displaymath}
	%	\Pif(h) = \left\{\bP \in \mathbb{R}_+^{n \times n}: \bP \mathbf{1}_n = \bh, \mathbf{G} \bP^T \mathbf{1}_n = \bzero_{\dF} \right\}.
	%\end{displaymath}
\end{definition}

The use of an entropy term in Eq.~(\ref{eq:otfe}) requires some justification, since we do not assume that $\bh$ represents a normalized probability distribution (i.e. sums to one). Therefore, $\bP$ also does not represent a normalized joint distribution. However, the $H(\bP)$ term still provides the practical advantage that it only admits couplings with $\bP_{ij} > 0$, which is why we require that $\bh_i > 0$. Furthermore, the addition of $H(\bP)$ makes the objective cost strongly convex.

%Since $\bh_i \leq 1$, we thus also have that $\bP_{ij} \leq 1$. 

%The use of the entropy term $H(\bP)$ in the objective Eq.~(\ref{eq:otfe}) requires some justification, since in contrast to the setting where the entropy regularization term is commonly used, we do not assume $\bh$ to represent a normalized probability distribution (i.e. sum to one). Therefore, $\bP$ also does not represent a normalized joint distribution. However, the $H(\bP)$ term maintains the practical advantage that it only admits couplings with $\bP_{ij} > 0$, which is why we require that $\bh_i > 0$. Since $\bh_i \leq 1$, we thus also have that $\bP_{ij} \leq 1$.

\subsubsection{Duality}
%The advantage of the entropic regularization is that the optimal coupling is no longer sparse. The problem can still be solved efficiently, however, because 
The OTF problem for a linear fairness notion has $n + \dF$ constraints, which is usually far fewer in number than the $n^2$ variables that make up the couplings in the primal problem. Because $\OTFe$ is also strongly convex, we will derive and then maximize the dual function instead.

\begin{propo}\label{prop:dual}
If $\bG$ has a non-empty null-space (i.e. $\exists \bff \in \mathbb{R}^n: \bff \neq \bzero_n \wedge \bG \bff = \bzero_{\dF}$), then the $\OTFe(h)$ problem has a unique solution for any $h: \mathcal{X} \to (0, 1]$. Moreover, the $\OTFe$ problem is strongly convex and then enjoys strong duality. Its dual function is
\begin{equation}\label{eq:dual_f}
	L(\blambda, \bmu) = \sum_i \blambda_i \bh_i - \epsilon \sum_{ij} \exp\left(\frac{1}{\epsilon} \left[-\bC_{ij} + \blambda_i + \sum_c \bmu_c \bG_{cj}\right]\right)
\end{equation}
with $\blambda \in \mathbb{R}^n$ and $\bmu \in \mathbb{R}^{\dF}$ the dual variable vectors for the marginalization and fairness constraints. 

%The values of the coupling are given by 
%\begin{equation}\label{eq:pij_expr}
%\bP^*_{ij}(\blambda, \bmu) = \exp\left(\frac{1}{\epsilon} \left[-\bC_{ij} + \blambda_i + \sum_c \bmu_c \bG_{cj}\right]\right).
%\end{equation}
\end{propo}

%Due to the strong duality, the values of the optimal coupling in $\OTFe$ are given by $\bP^*_{ij}(\blambda^*, \bmu^*)$, with the optimal dual variables $(\blambda^*, \bmu^*) = \argmax_{(\blambda, \bmu)} L(\blambda, \bmu)$.

\subsubsection{Optimization}\label{sec:optim}
Though Eq. (\ref{eq:dual_f}) could be maximized directly, we follow standard OT approaches \cite{peyreComputationalOptimalTransport2020} and perform our optimization with \textit{exact} coordinate ascent. This strategy is particularly useful here, because the $\blambda_i$ variable that maximizes $L(\blambda, \bmu)$ while $\bmu$ is fixed, is found independently of other variables in $\blambda$. All variables in $\blambda$ can therefore be updated at the same time:
\begin{equation}\label{eq:lambda_update}
	\blambda_i \leftarrow \epsilon \log \bh_i - \epsilon \log \sum_j \exp \left(\frac{1}{\epsilon} \left[- \bC_{ij} + \sum_c \bmu_c \bG_{cj}\right] \right)
\end{equation}
where we can use the stabilized log-sum-exp operation.

Unfortunately, there is no closed form expression for the $\bmu_c$ that maximizes $L(\blambda, \bmu)$. Instead, we preprocess $L(\blambda, \bmu)$ and numerically solve for each $\bmu_c$:
\begin{equation}\label{eq:mu_update}
	\bmu_c \leftarrow \argmin_{\bmu_c} \; \sum_{j} \bmeta_j(\blambda) \exp\left(\frac{1}{\epsilon} \sum_{k \neq c} \bmu_k \bG_{kj}\right) \exp\left(\frac{1}{\epsilon} \bmu_c \bG_{cj}\right)
\end{equation}
with $\bmeta_j(\blambda) = \sum_i \exp\left(\frac{1}{\epsilon} \left[- \bC_{ij} + \blambda_i\right]\right)$.

\subsubsection{Complexity}\label{sec:complexity}
The dual problem involves two variable vectors: $\blambda$ with length $n$ the number of samples and $\bmu$ with length $\dF$ the number of linear fairness constraints. It is commonly the case that $\dF \ll n$, because the number of distinguished protected groups is limited. This is in contrast with the traditional $\OTe$ problem, where the dual problem involves two dual variable vectors of length $n$.

%The dual function in Eq. (\ref{eq:dual_f}) involves two variable vectors. First, $\blambda$ with length $n$, the dataset size that $\OTFe$ is computed for. Second, $\bmu$ with length $\dF$, the number of linear fairness constraints. It is often the case that $\dF \ll n$, because even when multiple sensitive features are composed, each is either categorical with a limited number of possible values (i.e. distinguished groups) or continuous and then only has a dimensionality of 1 in our setting. This is in contrast with the traditional $\OTe$ problem, where the dual problem involves two dual variable vectors of length $n$.

In terms of computational complexity, the update of the full $\blambda$ has complexity $O(n(n + \dF))$. However, we stress that each $\blambda_i$ can be updated in parallel with complexity $O(n + \dF)$. When keeping the $\blambda$ vector fixed, we can perform a $O(n^2)$ operation to precompute the $\bmeta_j(\blambda)$ values for the $\bmu$ updates. Though these updates no closed form, the inner loop of the update for $\mu_c$ only has complexity complexity $O(n + \dF)$ for each inner loop, and should only be performed for $\dF$ variables. %Under the assumption that the computational complexity of a full maximization is far less than the $O(n^2)$ need to precompute $\bmeta$, the update of all $\bmu$ will not have a greater order of computational complexity than $\blambda$. 
By permitting a high tolerance on the convergence of $\blambda$, the necessary number of updates is also limited. In our experiments, we found that a single update is often enough.

We can achieve a memory complexity of $O(n + \dF)$. However, we can perform efficient matrix computations by storing the cost matrix $C$ and the fairness constraints matrix $G$. The memory required for practical use is therefore $O(n(n + \dF))$.

\subsection{Adjusted $\OTFe$}\label{sec:otf_adjusted}
A cause for concern with the smooth $\OTFe$ cost is that, as opposed to the standard $\OTF$, it does not necessarily equal zero if $h$ is already a fair score function. Indeed, while it may be feasible to achieve $\langle \bC, \bP \rangle = 0$, it is generally the case that the entropy $H(\bP) > 0$. We would thus like to find an interesting lower bound on $\OTFe$ that is tight when $h \in \mathcal{F}$.

\begin{definition}[$\OTFRe$]\label{def:otfr}
	The $\OTFe$ cost in Def.~\ref{def:otfe} is relaxed to the $\OTFRe$ cost as
	\begin{equation}\label{eq:otfre}
		\OTFRe(h) = \min_{\bP \in \Pif(h)} \langle \bC, \bP \rangle - \epsilon H(\bP)
	\end{equation}
	\begin{nospaceflalign*}
		\text{with} && \Pif(h) = \left\{\bP \in \mathbb{R}_+^{n \times n}: \bP \mathbf{1}_n = \bh, \card{\mathbf{G} \bP^T \mathbf{1}_n} \leq \card{\bG \bh} \right\}.&&	
	\end{nospaceflalign*}
\end{definition}

\begin{definition}[$\tOTFe$]\label{def:adj_otf}
	For a score function $h: \mathcal{X} \to (0, 1]$, the \textit{adjusted $\OTFe$ cost} is computed as
	\begin{equation}
		\tOTFe(h) = \OTFe(h) - \OTFRe(h).
	\end{equation}
\end{definition}

Whereas the $\OTFe(h)$ problem involved transport to fair score vectors ($\mathbf{G} \bP^T \mathbf{1}_n = \bzero_{\dF}$), the $\OTFRe(h)$ problem relaxes this constraint by allowing unfairness up to the unfairness already present in $h$. It is thus easily seen that $\OTFe(h) \geq \OTFRe(h)$. Interestingly, because this relaxation is bounded by the unfairness of $h$, we have that $\OTFRe(h)$ is a tight lower bound if $h$ is already fair.

\begin{propo}\label{prop:fair_zero}
	$h \in \mathcal{F} \implies \tOTFe(h) = 0.$
\end{propo}
In Appendix~\ref{sec:post},
we provide empirical intuition for this theoretical result by using $\tOTFe(h)$ as a postprocessing approach and illustrate that $\OTFe(h)$ gradually converges to $\OTFRe(h)$ as the fairness of $h$ improves.

%Proposition~\ref{prop:symm} allows us to solve the dual problem as was done in Sec.~\ref{sec:entr_reg}, yet this time without the $\bmu$ variables. The remaining dual variables for the constraint over the row marginals, now referred to as $\bkappa$ to avoid confusion, could already be computed independently from each other. Consequently, the optimal $\bkappa^*$ values that maximize the dual function $L(\bkappa)$ can now be computed directly:
%\begin{equation}\label{eq:kappa_update}
%	\bkappa^*_i =  \epsilon \log \bh_i - \epsilon \log \sum_j \exp \left(\frac{- \bC_{ij}}{\epsilon} \right).
%\end{equation}

The adjustment term $\OTFRe$ can be computed by using strong duality in a similar way as $\OTFe(h)$, though now with two dual variables $\bphi \in \mathbb{R}^{\dF}$ and $\bpsi \in \mathbb{R}^{\dF}$ that enforce the fairness inequality, in addition to the dual variable vector, here denoted by $\bkappa \in \mathbb{R}^{n}$, that enforces the row constraint on the coupling. The dual function of $\OTFRe$ is  
\begin{equation*}
	L(\bkappa, \bphi, \bpsi) = \sum_i \bkappa_i \bh_i + \sum_c\bgamma_c(\bphi_c + \bpsi_c) - \epsilon \sum_{ij} \exp\left(\frac{1}{\epsilon} \left[-\bC_{ij} + \bkappa_i + \sum_c (\bphi_c - \bpsi_c) \bG_{jc}\right]\right)
\end{equation*}
with $\bphi_c \leq 0$, $\bpsi_c \leq 0$ and $\bgamma_c = \card{\sum_j \bG_{cj} \bh_j}$. For a full discussion, we refer to Appendix~\ref{sec:otfr_comp}.

\subsection{Minimising Unfairness}\label{sec:min_otf}
Given an efficient way to compute the parameters of the adjusted, smooth $\tOTFe(h)$ cost, we can now jointly minimize the weighted sum of the classifier's loss and unfairness cost with respect to $h$.

Let $\mathcal{L}_Y(h)$ denote the cross-entropy loss of $h(X)$ for the output labels $Y$. To add our $\tOTFe(h)$ cost to this objective, we can pose the following optimization problem:
\begin{equation}\label{eq:joint_opt}
	\min_h (1 - \alpha) \mathcal{L}_Y(h) + \alpha \tOTFe(h)
\end{equation}
with $0 \leq \alpha \leq 1$ a hyperparameter that regulates the importance of the unfairness cost term $\tOTFe(h)$.

So far, we considered the scores vector of $h$ as constant with respect to the $\tOTFe(h)$ problem. However, the solutions for $\OTFe(h)$ and $\OTFRe(h)$ clearly depend on $h$. Let $\blambda^*$, $\bmu^*$, $\bkappa^*$, $\bphi^*$ and $\bpsi^*$ denote the fully converged dual variables. Then
\begin{equation}\label{eq:totfe_deriv}
	\frac{\partial}{\partial h} \tOTFe(h) = \frac{\partial}{\partial h} \left(L(\blambda^*, \bmu^*) - L(\bkappa^*, \bphi^*, \bpsi^*)\right)
\end{equation}
%	\begin{nospaceflalign*}
%	\text{with} && \blambda^*_i - \bkappa_i^* = \epsilon\log \sum_j \exp \left(\frac{- \bC_{ij}}{\epsilon} \right) - \epsilon\log \sum_j \exp \left(\frac{- \bC_{ij} + \sum_c \bmu_c^* \bG_{cj}}{\epsilon} \right).&&
%\end{nospaceflalign*}

%Observe that, though the update equations of $\blambda^*_i$ and $\bkappa_i^*$ depend on $\bh_i$, these terms cancel out in $\blambda^*_i - \bkappa_i^*$. Hence, the gradient of $\tOTFe(h)$ is linear. 
Even if we did not update the dual variables until convergence, we can still use intermediate values to approximate the true gradient \cite{genevayLearningGenerativeModels2018, feydyInterpolatingOptimalTransport2019}. %TODO put new line.
We further approximate the true $\tOTFe(h)$ cost by only computing it for a subset of the dataset, e.g. when batching is already used to optimize $\mathcal{L}_Y(h)$. The $\tOTFe(h)$ can be optimized on a batch by only using the values of $\bC$ and the columns of $\bG$ that relate to these data points. %Note: next line was added later.
For large datasets (including those we use in experiments), it is then unnecessary to precompute the $\bC$ matrix between all pairs of data points, since most pairs will never actually co-occur in randomly sampled batches.

%However, for small batch sizes this may lead to an incorrect estimation of the unfairness in the data.
%
%
%%%%%%%%%%%%%%%%%%%%%%%%%%%%%%%%%%%%%%%%%%%%%%%%%%%%%%%%%%%%
\section{Related Work}\label{sec:related}
%%%%%%%%%%%%%%%%%%%%%%%%%%%%%%%%%%%%%%%%%%%%%%%%%%%%%%%%%%%%
Fairness in machine learning is commonly understood in two (extreme) ways \cite{mehrabininarehSurveyBiasFairness2021}. First, the notion of \textit{individual} fairness states that similar individuals should be treated similarly \cite{dworkFairnessAwareness2012}. Second, \textit{group} fairness requires that as a group, people are treated equally. The primary goal of our work is to achieve group fairness by measuring the distance of a score function to a set of group-fair functions. Yet, to an extent we also approach individual fairness because our unfairness quantification takes the features of each individual into account in its cost function $c$. By appropriately choosing $c$, it may be possible to further improve upon individual fairness in future work.

Methods that aim to improve fairness can be categorized into three types: those that perform \textit{preprocessing} where the data is made more fair \cite{kamiranDataPreprocessingTechniques2012, calmonOptimizedPreProcessingDiscrimination2017}, as opposed to those that do \textit{postprocessing} where the model's predictions are modified to fit a notion of fairness \cite{hardtEqualityOpportunitySupervised2016}. As a fairness regularization term, our work clearly fits in the class of \textit{inprocessing} methods, i.e where the machine learning model or its training is directly altered to reduce unfairness \cite{caldersThreeNaiveBayes2010, zafarFairnessConstraintsFlexible2019, woodworthLearningNonDiscriminatoryPredictors2017, agarwalReductionsApproachFair2018, celisClassificationFairnessConstraints2019}. However, we highlight that the OTF cost could also be used as a postprocessing approach by setting $\alpha = 1$.
 %TODO more sources

Our work draws inspiration from two kinds of inprocessing methods. First, with those methods where a (still unfair) model is explicitly projected unto the set of fair models. A popular definition of distance for this projection is the Kullback-Leibler divergence \cite{alghamdiModelProjectionTheory2020, jiangIdentifyingCorrectingLabel2020, buylKLDivergenceGraphModel2021}. Second, our work uses Optimal Transport (OT) \cite{peyreComputationalOptimalTransport2020} theory. Several works have applied OT to fairness, though they almost all do so by finding barycenters \cite{silviaGeneralApproachFairness2020}: measures that are `equally far' from the measures belonging to each demographic. This type of method has been applied to preprocessing \cite{gordalizaObtainingFairnessUsing2019}, classification \cite{jiangIdentifyingCorrectingLabel2020, zehlikeMatchingCodeLaw2020} and regression \cite{chzhenFairRegressionWasserstein2020, gouicProjectionFairnessStatistical2020}, with the overarching goal being to move each group's measure closer to the barycenter. Such methods require fairness notions to be encoded in the format of explicit equality, between a limited amount of groups. They do not allow for more complex constraints such as those involving continuous or multiple sensitive attributes, and make strong assumption that the cost function is a distance metric. 

Our proposed method instead combines the idea of projecting the model to a fairness-constrained set and the use of OT to measure the cost of this projection. In this manner, our work is most similar to \cite{siTestingGroupFairness2021}, which computes the OT projection onto such a fair set. However, they only use the method as a group fairness test. Our method has the advantage that it computes a differentiable cost that can be added to the loss of any probabilistic classifier.

%
%
%%%%%%%%%%%%%%%%%%%%%%%%%%%%%%%%%%%%%%%%%%%%%%%%%%%%%%%%%%%%
\section{Experiments}\label{sec:experiments}
%%%%%%%%%%%%%%%%%%%%%%%%%%%%%%%%%%%%%%%%%%%%%%%%%%%%%%%%%%%%
%\subsection{Evaluation Setup}\label{sec:eval}
Our experiments were performed on four datasets from fair binary classification literature \cite{duaUCIMachineLearning2017, lequySurveyDatasetsFairnessaware2022} with different kinds of sensitive information. First, the UCI \textit{Adult} Census Income dataset with two binary sensitive attributes \textsc{sex} and \textsc{race}. Second, the UCI \textit{Bank} Marketing dataset, where we measure fairness with respect to the original continuous \textsc{age} values and the binary, quantized version \textsc{age\_binned}. Third, the \textit{Dutch} Census dataset with sensitive attribute \textsc{sex}. Fourth, the \textit{Diabetes} dataset with sensitive attribute \textsc{gender}. Additional information on datasets and hyperparameters is given in Appendix~\ref{sec:data_suppl} and \ref{sec:hyper} respectively.

%Before analyzing the results in Sec. \ref{sec:results}, we discuss the evaluated methods in Sec. \ref{sec:methods}.

\subsection{Methods}\label{sec:methods}
All experiments were performed using a logistic regression classifier. To achieve fairness, we jointly optimize fairness regularizers with the cross-entropy loss as in Eq. (\ref{eq:joint_opt}), and compute the gradient of their sum with respect to the parameters of the classifier.
%To achieve fairness, we evaluate three different fairness cost terms that are jointly optimized with the cross-entropy loss as in Eq. (\ref{eq:joint_opt}). These cost terms are referred to as \textit{OTF}, \textit{Norm} and \textit{Barycenter}. The method optimized without such a cost term is referred to as \textit{Unfair}. 

The OTF cost is the adjusted version from Def.~\ref{def:adj_otf}, with $\epsilon = 10^{-3}$ (different choices for $\epsilon$ are illustrated in Appendix~\ref{sec:eps_disc}). For the cost function $c$, we use the Euclidean distance between non-protected features. The fairness constraints for both sensitive features in the Adult dataset were combined by concatenating their $\bG$ matrices. %For the Bank dataset two OTF costs were computed: one using the continuous \textsc{age} values and one using the binned \textsc{age\_binned} attribute. The latter method is referred to as \textit{OTF (binned)}.

As a first baseline, we use the $L^1$ norm of the PDP and PEO fairness constraints and will refer to this method as the \textit{Norm} cost. This simple type of fairness regularizer is frequently proposed in literature \cite{kamishimaFairnessawareLearningRegularization2011, zafarFairnessConstraintsFlexible2019, wickUnlockingFairnessTradeoff2019, padalaFNNCAchievingFairness2021}. In our experiments, it thus represents the class of methods that can accommodate a flexible range of fairness constraints while only considering the output of a classifier and not the input features.
%As a first baseline, we used the DP and EO cost terms from the \textit{Norm}\footnote{Retrieved from \url{https://github.com/wbawakate/fairtorch} (MIT License)} library. These costs are the square error of the fairness constraint. Thus, it implements the regularized optimization problem referred to in \cite{zafarFairnessConstraintsFlexible2019}. 
%For the Adult dataset, we sum over the norms of the constraints for both the \textsc{sex} and \textsc{race} attributes. 

Our second baseline is an OT \textit{Barycenter}\footnote{Retrieved from \url{https://github.com/deepmind/wasserstein_fairness} (Apache License 2.0)} approach \cite{jiangWassersteinFairClassification2020}. This method directly minimizes the OT cost between the score distributions of two sensitive groups. However, the implementation can only minimize PDP, and does not admit a composition of linear fairness notions or continuous attributes. We therefore only minimize unfairness with respect to \textsc{sex} and \textsc{age\_binned}. To show the drawbacks of this limitation, we also report PDP unfairness with respect to \textsc{race} and the continuous \textsc{age}. 

The third baseline is \textit{Unfair}, which was a classifier trained without a fairness cost term.

%Additional information on the hyperparameters is given in the Appendix.% \ref{sec:hyper}.

\subsection{Evaluation}\label{sec:eval}
To measure the predictive power of the methods, the ROC AUC score was measured. Violation of PDP is quantified as the absolute value of the maximal Pearson correlation between the scores $h(X)$ and each dimension of the sensitive attributes $S_k$. For PEO, the violation is computed as the maximal PDP for the data points of each one-hot encoded output label $Y_l$. These measures mirror the requirements in Eq. (\ref{eq:cov}) and Eq. (\ref{eq:cond_cov}) that the covariances ought to equal zero, though we choose the Pearson correlation instead such that the violation is normalized between 0 and 1. Thus we avoid that a method may improve on linear fairness notions by squeezing the variance of its scores. 

\begin{figure}
	\centering
	\begin{subfigure}[b]{\textwidth}
		\centering
		\includegraphics[width=\textwidth]{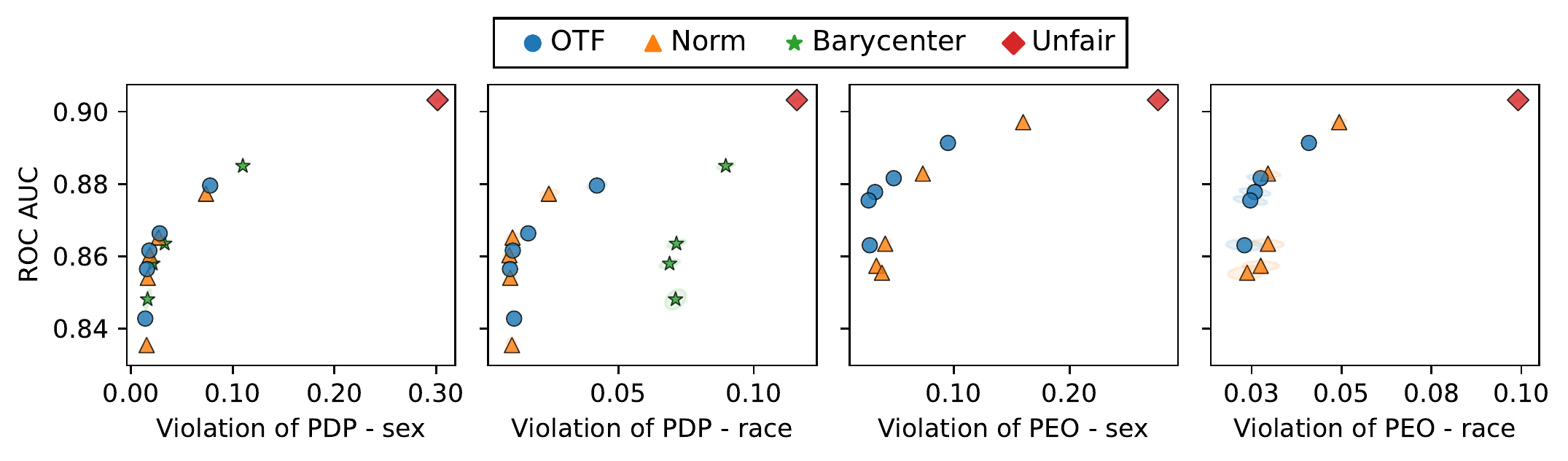}
		\caption{Adult dataset}
		\label{fig:adult}
	\end{subfigure}
	\hfill
	\begin{subfigure}[b]{\textwidth}
		\centering 
		\includegraphics[width=\textwidth]{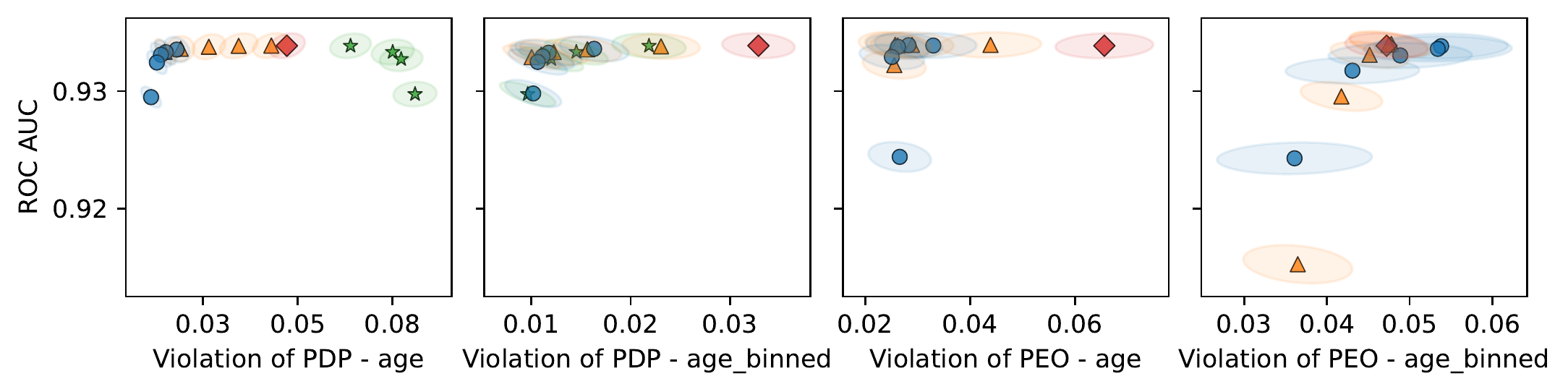}
		\caption{Bank dataset}
		\label{fig:bank}
	\end{subfigure}
	\begin{subfigure}[b]{0.495\textwidth}
	\centering 
	\includegraphics[width=\textwidth]{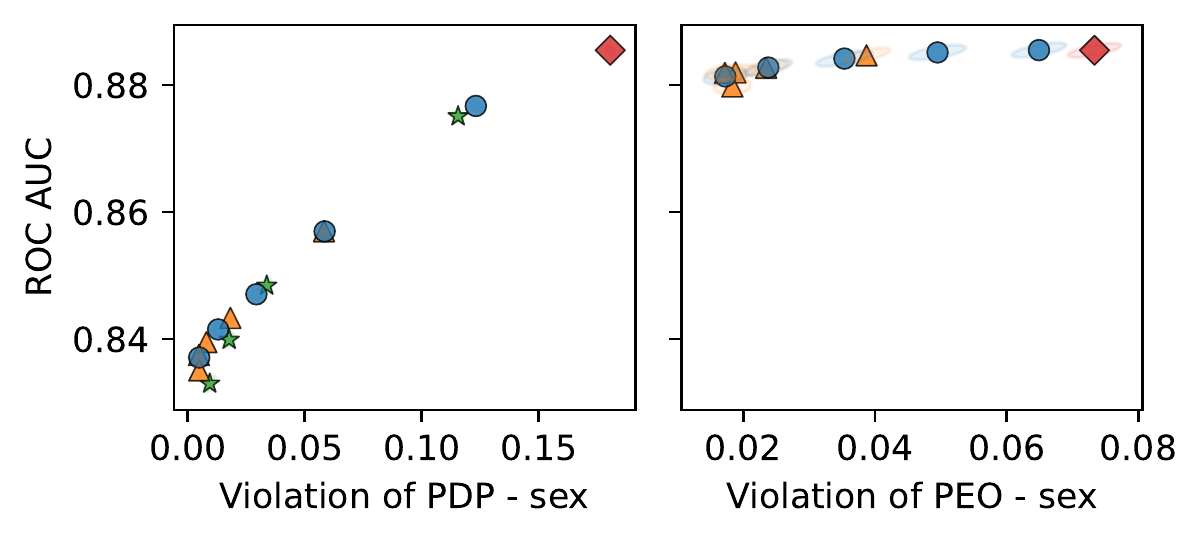}
	\caption{Dutch dataset}
	\label{fig:dutch}
	\end{subfigure}
	\begin{subfigure}[b]{0.495\textwidth}
	\centering 
	\includegraphics[width=\textwidth]{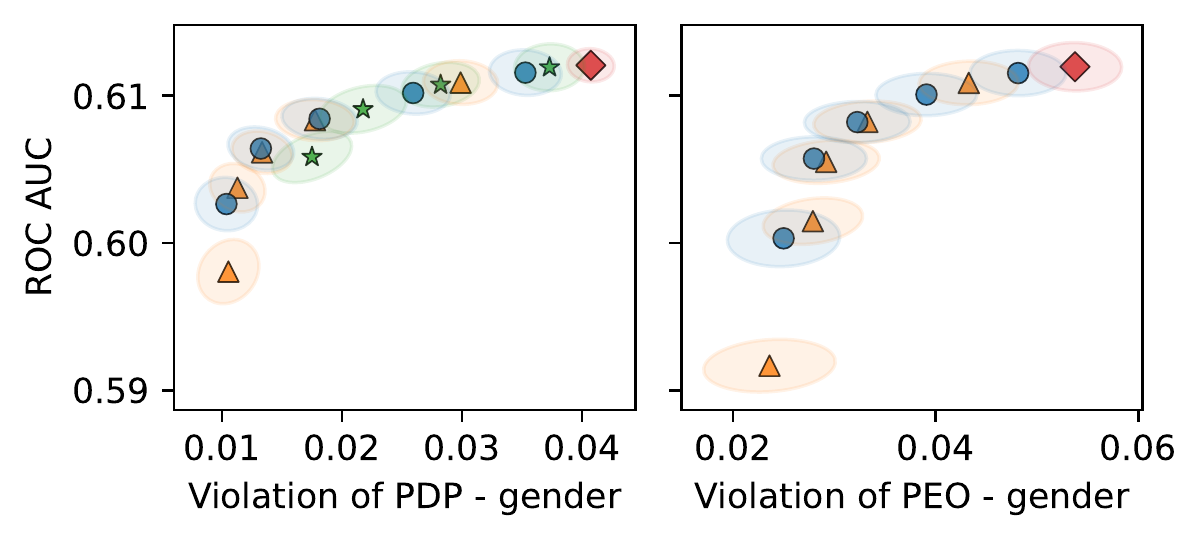}
	\caption{Diabetes dataset}
	\label{fig:diabetes}
	\end{subfigure}
	\caption{Test set results for the methods that were trained to reduce the evaluated fairness measure (PDP or PEO). Violation of PDP (and PEO) is computed as the maximal absolute Pearson correlation between the probability scores (conditioned on the output labels) and each sensitive attribute.}
	\label{fig:exps}
\end{figure}

All methods were tested for various strengths of $\alpha$ (see Eq. (\ref{eq:joint_opt})). Each configuration of each method (i.e. each $\alpha$ value and fairness notion) was tested for 10 train-test splits with proportions 80/20. We report mean test set scores in Fig.~\ref{fig:exps} and show the confidence ellipse\footnote{We use the variance of the estimator of the mean, which is smaller than that of individual scores} of their estimator for the first standard deviation. The train set results are reported in Appendix~\ref{sec:train_res} and show the same trends.

Note that for the OTF and Norm regularizers, we omit fairness scores for notions that were not optimized in that configuration. While \textsc{sex} and \textsc{race} are minimized at the same time for the Adult dataset, the \textsc{age} and \textsc{age\_binned} attributes were minimized in separate training runs.

All experiments were conducted using half the hyperthreads on an internal machine equipped with a 12 Core Intel(R) Xeon(R) Gold processor and 256 GB of RAM. In total, all experiments, including preliminary and failed experiments, used this machine for approximately 200 hours.

\subsection{Results}\label{sec:results}
On all datasets, our OTF method makes a similar trade-off between AUC and PDP fairness as the baselines for various $\alpha$ values. However, OTF clearly outperforms Norm for both sensitive attributes on the Adult dataset in Fig.~\ref{fig:adult} when they are trained to minimize PEO, especially in the lower violation range. Similarly, OTF achieves a better trade-off between AUC and PEO violation on the Diabetes dataset in Fig.~\ref{fig:diabetes}, especially on the train set (see Appendix~\ref{sec:train_res}). This advantage in minimizing PEO may be due to the fact that OTF has the incentive to achieve fairness by exchanging classifier scores between similar individuals. Conditioning on the output label rewards such exchanges, as individuals with similar labels can be expected to have similar features. 

Furthermore, we remark that our OTF method is as flexible as the Norm baseline in minimizing either PDP or PEO with respect to multiple sensitive attributes on the Adult dataset and the continuous \textsc{age} attribute on the Bank dataset in Fig~\ref{fig:bank}. Implementations of differentiable regularizers that can accommodate such attributes are scarcely available. For example, the Barycenter method was not trained to improve fairness with respect to \textsc{race}, as its implementation can not handle multiple sensitive attributes. Since it can only minimize the OT cost between a limited number of score distributions, it is naturally ill-equipped for continuous attributes. In separate runs on the Bank dataset, we therefore also trained OTF and Norm to minimize unfairness with respect to \textsc{age\_binned}. Though the results are subject to noise, we observe similar trends as for the other non-continuous attributes: our method is clearly on par with the baselines. On the train set results (see Appendix~\ref{sec:train_res}) the OTF method significantly outperforms the Norm baseline for PEO.

On the Adult, Bank and Diabetes datasets, the proposed OTF method thus displays its advantage in both an effective AUC-fairness trade-off, and a flexible capacity for fairness. However, we note that on the Dutch dataset in Fig.~\ref{fig:dutch}, no method seems to display a significantly better trade-off. We hypothesize that no such advantage is seen here because PEO, the fairness notion where our method usually achieves superior results, may be relatively easy to satisfy on this dataset.

\subsection{Limitations and Impact}\label{sec:limits}
We nuance our results by pointing out that the OTF method has a higher computation complexity ($O(n(n+\dF))$) than Norm, which has complexity $O(n\dF)$ with $n$ the batch size and $\dF$ the number of fairness constraints. Furthermore, as discussed in Sec.~\ref{sec:fair_class}, we only evaluate \textit{linear} fairness notions and leave an extension to non-linear notions of fairness \cite{celisClassificationFairnessConstraints2019} for future work. % we therefore miss important group fairness notions, and only evaluate a linear form of independence for continuous sensitive attributes. %Also, the group fairness definitions we employed may ignore discrimination with respect to structured subgroups \cite{kearnsPreventingFairnessGerrymandering2018}. 
 
The fairness of a system should always be judged based on a holistic consideration of the context of the system \cite{bennettWhatPointFairness2020} and relevant ethical principles \cite{binnsApparentConflictIndividual2020}. Without it, any method to improve fairness may reinforce the underlying injustice that led to the risks of discrimination that we aim to solve \cite{hoffmannWhereFairnessFails2019}.

%
%
%%%%%%%%%%%%%%%%%%%%%%%%%%%%%%%%%%%%%%%%%%%%%%%%%%%%%%%%%%%%
\section{Conclusion}
%%%%%%%%%%%%%%%%%%%%%%%%%%%%%%%%%%%%%%%%%%%%%%%%%%%%%%%%%%%%
In this paper, we constructed the Optimal Transport to Fairness (OTF) method as a differentiable fairness regularization term, which combines the advantages of flexible linear fairness constraints with Optimal Transport theory, thereby taking non-protected similarities between individuals into account. In experiments, we show that our final method achieves a similar AUC-fairness trade-off as other cost terms for notions of fairness inspired by demographic parity, and significantly better for notions inspired by equalized odds. At the same time, OTF empirically displays why its increased flexibility, i.e. its ability to improve all linear fairness notions with respect to multiple sensitive attributes that may be categorical or continuous, provides an advantage over previous fairness applications of OT. 

In the future, we hope to further investigate the properties of OT-based fairness, inspired by its clear advantage in achieving equalized odds. Though our method was only applied to linear fairness notions, OT may provide an opportunity to better achieve non-linear notions of fairness, e.g. through a creative choice of the cost function that directs the transport of classifier scores.

%%%%%%%%%%%%%%%%%%%%% ACKNOWLEDGEMENTS %%%%%%%%%%%%%%%%%%%%%%%%%%
\begin{ack}
The research leading to these results has received funding from the European Research Council under the European Union's Seventh Framework Programme (FP7/2007-2013) (ERC Grant Agreement no. 615517), and under the European Union’s Horizon 2020 research and innovation programme (ERC Grant Agreement no. 963924), from the Flemish Government under the ``Onderzoeksprogramma Artificiële Intelligentie (AI) Vlaanderen'' programme, and from the FWO (project no. G0F9816N, 3G042220). MB is supported by a doctoral scholarship from the Special Research Fund (BOF) of Ghent University (reference number: BOF20/DOC/144).
\end{ack}
%Use unnumbered first level headings for the acknowledgments. All acknowledgments go at the end of the paper before the list of references. Moreover, you are required to declare funding (financial activities supporting the submitted work) and competing interests (related financial activities outside the submitted work). More information about this disclosure can be found at: \url{https://neurips.cc/Conferences/2022/PaperInformation/FundingDisclosure}. Do {\bf not} include this section in the anonymized submission, only in the final paper. You can use the \texttt{ack} environment provided in the style file to autmoatically hide this section in the anonymized submission.
%\end{ack}

%%%%%%%%%%%%%%%%%%%%% REFERENCES %%%%%%%%%%%%%%%%%%%%%%%%%%
%TODO: parse references manually here. Just .bib?
%\section*{References}
%References follow the acknowledgments. Use unnumbered first-level heading for the references. Any choice of citation style is acceptable as long as you are consistent. It is permissible to reduce the font size to \verb+small+ (9 point) when listing the references. Note that the Reference section does not count towards the page limit. \medskip

%{
%	\small
%}
\bibliographystyle{plain}
\bibliography{references}

%%%%%%%%%%%%%%%%%%%%% CHECKLIST %%%%%%%%%%%%%%%%%%%%%%%%%%
%%%%%%%%%%%%%%%%%%%%%%%%%%%%%%%%%%%%%%%%%%%%%%%%%%%%%%%%%%%%
\section*{Checklist}

\begin{enumerate}

\item For all authors...
\begin{enumerate}
  \item Do the main claims made in the abstract and introduction accurately reflect the paper's contributions and scope?
    \answerYes{In Sec.~\ref{sec:otf}, we proposed the OTF cost term and deployed tricks to minimize it effectively. In Sec.~\ref{sec:experiments}, we show that the results of using our proposed method are close to the baselines or surpassing it.}
  \item Did you describe the limitations of your work?
    \answerYes{We describe limitations of our considered fairness notions in Sec.~\ref{sec:fair_class} and the complexity of our proposed OTF cost term in Sec.~\ref{sec:complexity}. We briefly discuss overall limitations of fairness methods in Sec.~\ref{sec:limits}.}
  \item Did you discuss any potential negative societal impacts of your work?
    \answerYes{We briefly discuss the danger that inconsiderately applied fairness methods can pose in Sec.~\ref{sec:limits}}
  \item Have you read the ethics review guidelines and ensured that your paper conforms to them?
    \answerYes{}
\end{enumerate}

\item If you are including theoretical results...
\begin{enumerate}
  \item Did you state the full set of assumptions of all theoretical results?
    \answerYes{Our method is constructed in several steps. When required, we discuss and add additional assumptions.}
        \item Did you include complete proofs of all theoretical results?
    \answerYes{Proofs of all propositions and derivations of other results are included in the Appendix.}
\end{enumerate}

\item If you ran experiments...
\begin{enumerate}
  \item Did you include the code, data, and instructions needed to reproduce the main experimental results (either in the supplemental material or as a URL)?
    \answerYes{There are references to the datasets we used in the Appendix. Code and a pipeline to run it are included in the supplemental material.}
  \item Did you specify all the training details (e.g., data splits, hyperparameters, how they were chosen)?
    \answerYes{Initial training details are given in Sec.\ref{sec:eval}. More details of the hyperparameters are included in the Appendix.}
        \item Did you report error bars (e.g., with respect to the random seed after running experiments multiple times)?
    \answerYes{Since we show the trade-off between two goals (accuracy and fairness) we show the confidence ellipses.}
        \item Did you include the total amount of compute and the type of resources used (e.g., type of GPUs, internal cluster, or cloud provider)?
    \answerYes{This is mentioned at the end of Sec.~\ref{sec:eval}.}
\end{enumerate}

\item If you are using existing assets (e.g., code, data, models) or curating/releasing new assets...
\begin{enumerate}
  \item If your work uses existing assets, did you cite the creators?
    \answerYes{For the Barycenter code, a reference to the Github implementation is included. For datasets, we refer to the UCI repository.}
  \item Did you mention the license of the assets?
    \answerYes{For the Barycenter code, we include the license. We refer to the data according to the requested citation policy.}
  \item Did you include any new assets either in the supplemental material or as a URL?
    \answerNA{}
  \item Did you discuss whether and how consent was obtained from people whose data you're using/curating?
    \answerNo{The datasets we use are popular, public datasets in machine learning fairness literature. Obtaining consent is infeasible. However, the data is highly anonymized and its popularity promotes reproducibility in the community. }
  \item Did you discuss whether the data you are using/curating contains personally identifiable information or offensive content?
    \answerYes{As briefly mentioned in the Appendix, the data is anonymous and free from any personal identifiers.}
\end{enumerate}

\item If you used crowdsourcing or conducted research with human subjects...
\begin{enumerate}
  \item Did you include the full text of instructions given to participants and screenshots, if applicable?
    \answerNA{}
  \item Did you describe any potential participant risks, with links to Institutional Review Board (IRB) approvals, if applicable?
    \answerNA{}
  \item Did you include the estimated hourly wage paid to participants and the total amount spent on participant compensation?
    \answerNA{}
\end{enumerate}

\end{enumerate}

%%%%%%%%%%%%%%%%%%%%% APPENDIX %%%%%%%%%%%%%%%%%%%%%%%%%%
\appendix

\section{Derivations and Proofs}\label{sec:proofs}
\subsection{Proof of Proposition 1}% dual function
With $\bh_i > 0$, we assume that the null-space of $\bG$ is non-empty, i.e.
\begin{equation*}
\exists \bff \in \mathbb{R}^n: \bff \neq \bzero_n \wedge \bG \bff = \bzero_{\dF}.
\end{equation*}

Then, $\bG \bff = \bzero_{\dF}$ implies that uniformly rescaled score vectors $s\bff$ with $s \in \mathbb{R}$ are also fair. One of those fair, rescaled score vectors will have the same total mass as $\bh$:
\begin{equation*}
\exists s \in \mathbb{R}: \bG (s\bff) = \bzero_{\dF} \wedge (s\bff)^T \mathbf{1}_n =  \bh^T \mathbf{1}_n.
\end{equation*}

Therefore, for such a $\bG$ there is always a fair score function that $h$ can be transported to (i.e. $\Pif(h) \neq \emptyset$). Such $\OTe$ problems always have a solution coupling \cite{peyreComputationalOptimalTransport2019}. Thus, there exists an optimal coupling for the $\OTFe$ problem.

The objective cost of $\OTFe$ is the strictly convex functional
\begin{equation*}
\langle \bC, \bP \rangle - \epsilon H(\bP)
\end{equation*}
and the equality constraints that define the valid set of couplings $\Pif(h)$ are affine. The $\OTFe$ problem is thus convex and enjoys strong duality. Due to the strict convexity of the objective cost, the optimal coupling solution is unique.

Recall that $H(\bP) = - \sum_{ij} \bP_{ij} \left(\log\left(\bP_{ij}\right) - 1\right)$. The Lagrangian of the $\OTFe$ problem is given by
\begin{equation*}
	\Lambda(\bP, \blambda, \bmu) = \sum_{ij} \bC_{ij}\bP_{ij} - \sum_i \blambda_i \left(\sum_j \bP_{ij} - \bh_i \right) - \sum_c \bmu_c \left(\sum_{ij} \bP_{ij} \bG_{cj}\right) + \epsilon \sum_{ij} \bP_{ij} \left(\log \bP_{ij} - 1\right)
\end{equation*}
where $\blambda \in \mathbb{R}^n$ and $\bmu \in \mathbb{R}^{\dF}$ denote the dual variable vectors for the marginalization and fairness constraints respectively.

The Lagrangian, which is continuously differentiable around the optimal coupling, is written as a sum over the elements of $\bP$. We can thus minimize $\Lambda(\bP, \blambda, \bmu)$ by setting the derivative $\frac{\partial \Lambda(\bP, \blambda, \bmu)}{\partial \bP_{ij}} = 0$:
\begin{align*}
	&\frac{\partial \Lambda(\bP, \blambda, \bmu)}{\partial \bP_{ij}} = \bC_{ij} - \blambda_i - \sum_c \bmu_c \bG_{cj} + \epsilon \log \bP_{ij} = 0\\
	&\implies \bP^*_{ij}(\blambda, \bmu) = \exp\left(\frac{1}{\epsilon} \left[-\bC_{ij} + \blambda_i + \sum_c \bmu_c \bG_{cj}\right]\right).
\end{align*}%\label{eq:pij_expr}

This results in the dual function
\begin{equation*}
	L(\blambda, \bmu) = \Lambda(\bP^*(\blambda, \bmu), \blambda, \bmu) = \sum_i \blambda_i \bh_i - \epsilon \sum_{ij} \exp\left(\frac{1}{\epsilon} \left[-\bC_{ij} + \blambda_i + \sum_c \bmu_c \bG_{cj}\right]\right).
\end{equation*}

Due to the strong duality, the values of the optimal coupling in $\OTFe$ are given by $\bP^*_{ij}(\blambda^*, \bmu^*)$, with the optimal dual variables $(\blambda^*, \bmu^*) = \argmax_{(\blambda, \bmu)} L(\blambda, \bmu)$.

\subsection{Derivation of the Update Eq. (9) and Eq. (10)}\label{sec:deriv_lambda_mu}
We aim to update the $\blambda_i$ and $\bmu_c$ variables through exact coordinate ascent of the dual function $L(\blambda, \bmu)$. To this end, we derive expressions for the univariate updates of $L(\blambda, \bmu)$. For $\blambda_i$, the update is given by
\begin{align*}
&\frac{\partial L(\blambda, \bmu)}{\partial \blambda_i} = \bh_i - \exp\left(\frac{\blambda_i}{\epsilon}\right)\sum_j \exp\left(\frac{1}{\epsilon} \left[-\bC_{ij} + \sum_c \bmu_c \bG_{cj}\right]\right) = 0\\
&\implies \blambda_i^*(\blambda_1, ..., \blambda_{i-1}, \blambda_{i+1}, ..., \blambda_n, \bmu) = \epsilon \log \bh_i - \epsilon \log \sum_j \exp \left(\frac{1}{\epsilon} \left[- \bC_{ij} + \sum_c \bmu_c \bG_{cj}\right] \right)
\end{align*}
where $\blambda_i^*(\blambda_1, ..., \blambda_{i-1}, \blambda_{i+1}, ..., \blambda_n, \bmu)$ denotes the value of $\blambda_i$ that maximizes $L(\blambda, \bmu)$. This update has the useful property that it is independent of other variables in $\lambda$, i.e. $\blambda_i^*(\blambda_1, ..., \blambda_{i-1}, \blambda_{i+1}, ..., \blambda_n, \bmu) = \blambda_i^*(\bmu)$

The update for $\bmu_c$ generally does not have a closed form expression, because it can not be isolated from the expression of the gradient:
\begin{equation*}
\frac{\partial L(\blambda, \bmu)}{\partial \bmu_c} = - \sum_j \bG_{cj} \bmeta_j(\blambda) \exp\left(\sum_{k \neq c} \frac{\bmu_k \bG_{kj}}{\epsilon} \right) \exp\left(\frac{\bmu_c \bG_{cj}}{\epsilon} \right)
\end{equation*}
with $\bmeta_j(\blambda) = \sum_i \exp\left(\frac{1}{\epsilon} \left[- \bC_{ij} + \blambda_i\right]\right)$.

Instead, we maximize the dual function $L(\blambda, \bmu)$ with respect to $\bmu_c$ numerically.
\begin{align*}
\bmu_c^*(\blambda, \bmu_1, ..., \bmu_{c-1}, \bmu_{c+1}, ..., \bmu_{\dF}) &= \argmax_{\bmu_c} L(\blambda, \bmu)\\
&= \argmax_{\bmu_c} \; -\epsilon \sum_{j} \bmeta_j(\blambda) \exp\left(\frac{1}{\epsilon} \sum_{k \neq c} \bmu_k \bG_{kj}\right) \exp\left(\frac{1}{\epsilon} \bmu_c \bG_{cj}\right)\\
&= \argmin_{\bmu_c} \; \sum_{j} \bmeta_j(\blambda) \exp\left(\frac{1}{\epsilon} \sum_{k \neq c} \bmu_k \bG_{kj}\right) \exp\left(\frac{1}{\epsilon} \bmu_c \bG_{cj}\right)
\end{align*}

Let $\blambda_i^{(t)}$ and $\bmu_c^{(t)}$ denote the variables at the end of iteration $t$. In every iteration $t$, we set
\begin{align*}
\forall i \in [n]:\; &\blambda_i^{(t)} \leftarrow \blambda_i^*(\blambda_1^{(t)}, ..., \blambda_{i-1}^{(t)}, \blambda_{i+1}^{(t-1)}, ..., \blambda_n^{(t-1)}, \bmu^{(t-1)})\\
\forall c \in [\dF]:\; &\bmu_c^{(t)} \leftarrow \bmu_c^*(\blambda^{(t)}, \bmu_1^{(t)}, ..., \bmu_{c-1}^{(t)}, \bmu_{c+1}^{(t-1)}, ..., \bmu_{\dF}^{(t-1)})
\end{align*}

\subsection{Proof of Proposition 2}
Per Definition 1, we have that $h \in \mathcal{F} \iff \bG\bh = \bzero_{\dF}$. In this case, the column constraints upon the set of valid couplings for $\OTFRe(h)$ are simplified as follows:
\begin{equation*}
	\card{\mathbf{G} \bP^T \mathbf{1}_n} \leq \card{\bG \bh} \iff \card{\mathbf{G} \bP^T \mathbf{1}_n} \leq \card{\bzero_{\dF}} \iff \mathbf{G} \bP^T \mathbf{1}_n = \bzero_{\dF}
\end{equation*}

These simplified column constraints are equal to those posed upon couplings for $\OTFe(h)$. Since the other constraints and objective cost were already the same, we thus have that
\begin{equation*}
h \in \mathcal{F} \implies \OTFe(h) = \OTFRe(h) 
\end{equation*}

Thus, for $h \in \mathcal{F}$, the adjusted $\tOTFe(h)$ cost is 
\begin{equation*}
\tOTFe(h) = \OTFe(h) - \OTFRe(h) = 0.
\end{equation*}

We leave a study of the assumptions needed for the opposite implication $\left(\tOTFe(h) = 0 \stackrel{?}{\implies} h \in \mathcal{F}\right)$ for future work. For now, we observe that $\OTFe(h) = \OTFRe(h)$ is possible for a non-fair $h \notin \mathcal{F}$ if the optimal coupling of the relaxed $\OTFRe(h)$ cost coincidentally happens to transport $\bh$ to a score vector that is fair. %For example, if $\forall \bC_{ij} = 0$, then the optimal coupling for $\OTFRe(h)$ is found by only maximizing its element-wise entropy. For certain $\bh$, 

\subsection{Computation of $\OTFRe(h)$}\label{sec:otfr_comp}
The only difference between the $\OTFe$ and $\OTFRe$ problems is in the column constraints posed upon the couplings. The equality constraint in $\OTFe$ (i.e. $\mathbf{G} \bP^T \mathbf{1}_n = \bzero_{\dF}$) is relaxed to bounds on the unfairness of $h$ in $\OTFRe$ (i.e. $\card{\mathbf{G} \bP^T \mathbf{1}_n} \leq \card{\bG \bh}$). Because the objective of $\OTFRe$ remains strongly convex, and the relaxed constraints are still affine, we maintain the existence, uniqueness and strong duality properties from Proposition 1. Similarly to $\OTFe$, we thus solve the dual problem for $\OTFRe$.

For $\OTFRe$, we write out the Lagrangian as follows:
\begin{align*}
	\Lambda(\bP, \bkappa, \bphi, \bpsi) = &\sum_{ij} \bC_{ij}\bP_{ij} + \epsilon \sum_{ij} \bP_{ij} \left(\log \bP_{ij} - 1\right) - \sum_i \bkappa_i \left(\sum_j \bP_{ij} - \bh_i \right)\\ &- \sum_c \bphi_c \left(\sum_{ij} \bP_{ij} \bG_{cj} - \bgamma_c \right) - \sum_c \bpsi_c \left(-\sum_{ij} \bP_{ij} \bG_{cj} - \bgamma_c \right) 
\end{align*}
with $\bkappa \in \mathbb{R}^n$ the dual variable vector for the row constraints and $\bphi \in \mathbb{R}^\dF$ and $\bpsi \in \mathbb{R}^\dF$ the dual variable vectors for the fairness bounds, where we require that $\bphi_c < 0$ and $\bpsi_c < 0$. We also use $\bgamma_c = \card{\sum_j \bG_{cj} \bh_j}$ to simplify notation.

We minimize $\Lambda(\bP, \bkappa, \bphi, \bpsi)$ by setting the derivative $\frac{\partial \Lambda(\bP, \bkappa, \bphi, \bpsi)}{\partial \bP_{ij}} = 0$:
\begin{align*}
	&\frac{\partial \Lambda(\bP, \bkappa, \bphi, \bpsi)}{\partial \bP_{ij}} = \bC_{ij} - \bkappa_i - \sum_c (\bphi_c - \bpsi_c) \bG_{cj} + \epsilon \log \bP_{ij} = 0\\
	&\implies \bP^*_{ij}(\bkappa, \bphi) = \exp\left(\frac{1}{\epsilon} \left[-\bC_{ij} + \bkappa_i + \sum_c (\bphi_c - \bpsi_c) \bG_{cj}\right]\right).
\end{align*}%\label{eq:pij_expr}

This results in the dual function $\Lambda(\bP^*(\bkappa, \bphi, \bpsi), \bkappa, \bphi, \bpsi) = L(\bkappa, \bphi, \bpsi)$
\begin{equation*}
	L(\bkappa, \bphi, \bpsi) = \sum_i \bkappa_i \bh_i + \sum_c(\bphi_c + \bpsi_c)\bgamma_c - \epsilon \sum_{ij} \exp\left(\frac{1}{\epsilon} \left[-\bC_{ij} + \bkappa_i + \sum_c (\bphi_c - \bpsi_c) \bG_{cj}\right]\right).
\end{equation*}

We again maximize $L(\bkappa, \bphi, \bpsi)$ through the exact coordinate ascent scheme described in Appendix \ref{sec:deriv_lambda_mu}. The update equations are as follows:
\begin{align*}
\bkappa_i^* &\leftarrow \epsilon \log \bh_i - \epsilon \log \sum_j \exp \left(\frac{1}{\epsilon} \left[- \bC_{ij} + \sum_c (\bphi_c - \bpsi_c) \bG_{cj}\right] \right)\\
\bphi_c^* &\leftarrow \argmax_{\bphi_c} \; \bgamma_c \bphi_c - \epsilon \sum_{j} \bmeta_j(\bkappa) \exp\left(\frac{1}{\epsilon} \sum_k (\bphi_k - \bpsi_k) \bG_{kj}\right)\\
\bpsi_c^* &\leftarrow \argmax_{\bpsi_c} \; \bgamma_c \bpsi_c - \epsilon \sum_{j} \bmeta_j(\bkappa) \exp\left(\frac{1}{\epsilon} \sum_k (\bphi_k - \bpsi_k) \bG_{kj}\right)
\end{align*}
where we again note the possibility to precompute $\bmeta_j(\bkappa) = \sum_i \exp\left(\frac{1}{\epsilon} \left[- \bC_{ij} + \bkappa_i\right]\right)$.

\section{Additional Experiment Results}
\subsection{Post-Processing using $\tOTFe(h)$}\label{sec:post}
\begin{figure}[h]
	\centering
	\includegraphics[width=0.7\textwidth]{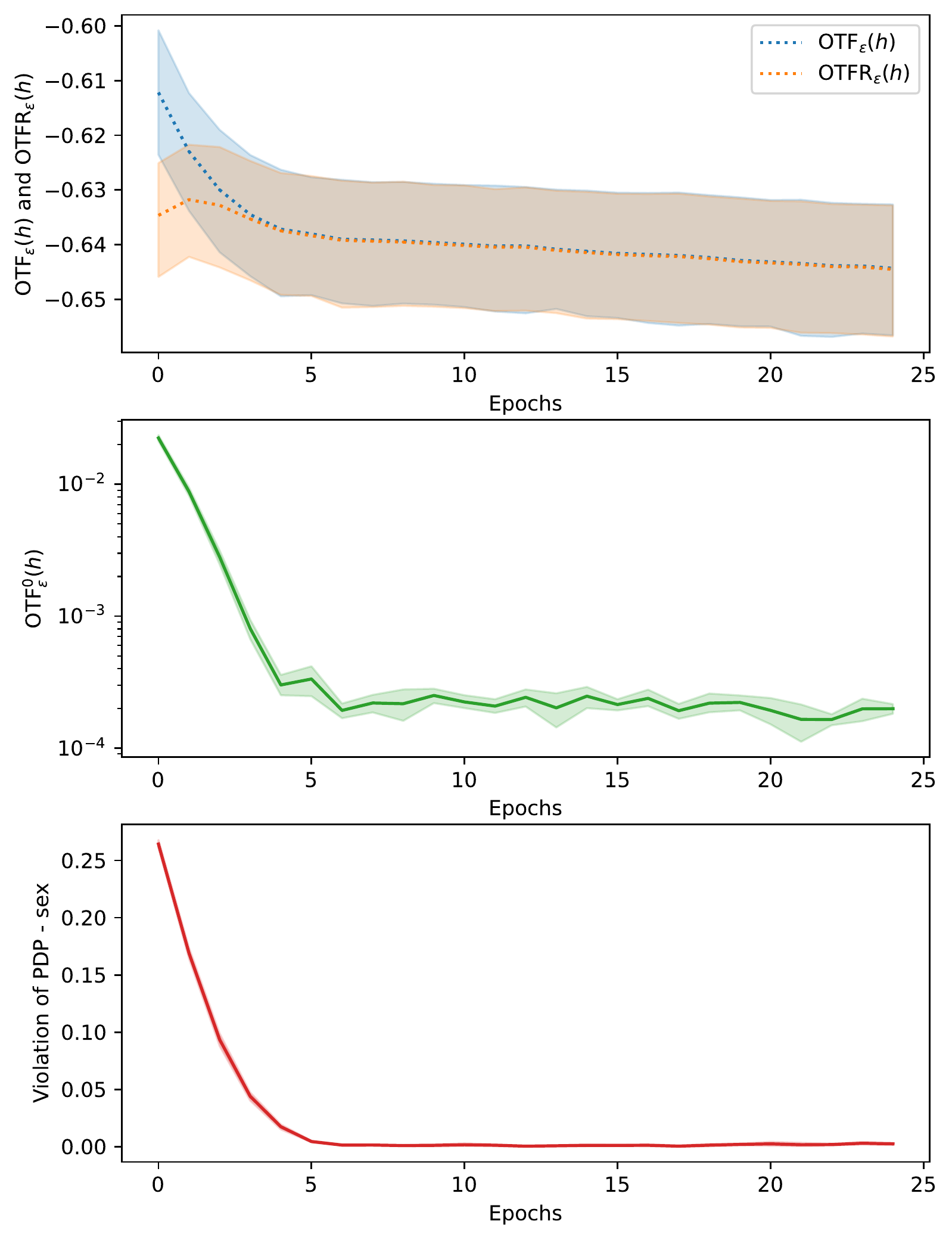}
	\caption{For 25 epochs of post-processing of a trained classifier, these plots show the trends of the $\OTFe(h)$ and $\OTFRe(h)$ terms, their gap defined as $\tOTFe(h)$, and the violation of the PDP fairness constraint with respect to the \textsc{sex} attribute. Since the experiment was repeated five times, we show the mean curves and the confidence interval for the first standard deviation.} 
	\label{fig:proj}
\end{figure}

Though we jointly minimize $\mathcal{L}_Y(h)$ and $\tOTFe(h)$ in our main experiments, we visualize the use of $\tOTFe(h)$ as a post-processing approach in Fig.~\ref{fig:proj}. This was done by first training the logistic regression classifier $h$ on the Adult dataset for 25 epochs by only minimizing cross-entropy (i.e. with $\alpha = 0$), and afterwards minimizing only the adjusted $\tOTFe$ cost for 25 epochs with $\alpha = 1$ and $\epsilon = 10^{-3}$. Here, the violation of the PDP fairness constraint with respect to only the \textsc{sex} attribute was minimized. We average out over five such runs with different random seeds and train-test splits.

We observe that $\OTFe(h)$ and $\OTFRe(h)$ are indeed not zero, because their optimizations involve maximizing the entropy term $H(\bP)$. However, as shown by the trajectory of the $\tOTFe(h)$ curve, the gap between them exponentially decreases from approximately $10^{-2}$ to $10^{-4}$. As their gap decreases, we also see that the PDP violation, measured as the maximal Pearson correlation discussed in Sec.~\ref{sec:eval}, trends towards zero. 

\subsection{Impact of Smoothing Factor $\epsilon$}\label{sec:eps_disc}
\begin{figure}
\begin{tabular}{lccccc}
%& $\alpha = 0.1$
 & $\alpha = 0.1$ & $\alpha = 0.3$ & $\alpha = 0.5$ & $\alpha = 0.7$ & $\alpha = 0.9$\\
$\epsilon = 0.001$ & 
\distrimage{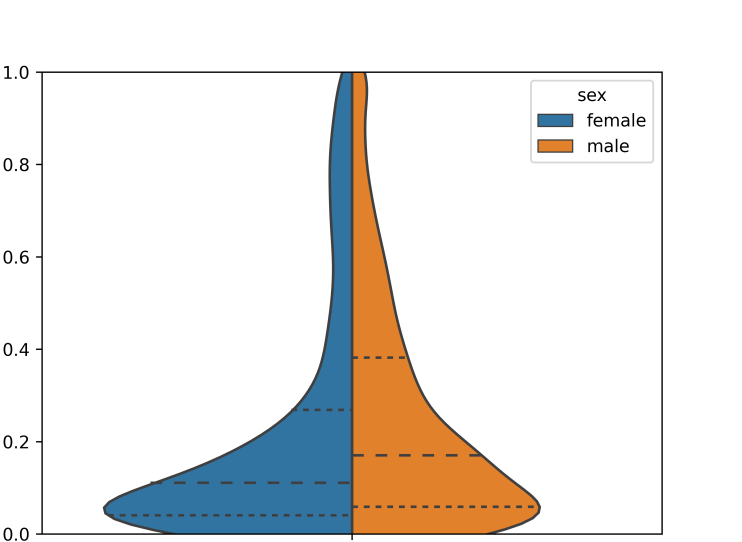} &
\distrimage{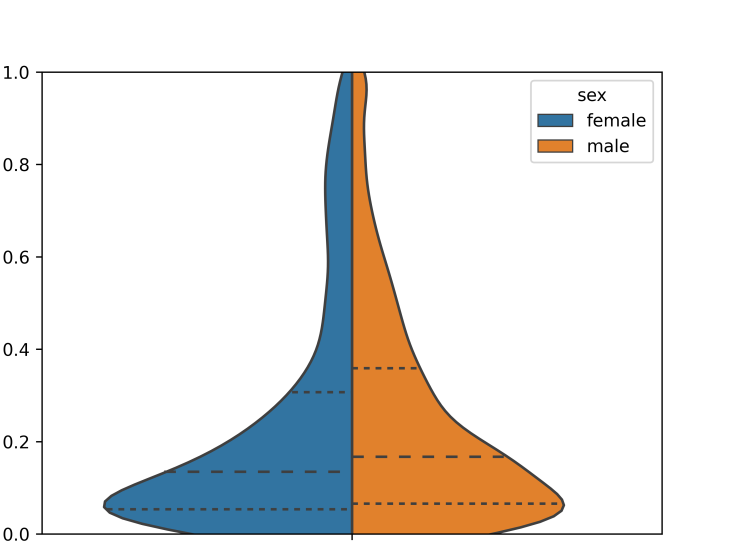} &
\distrimage{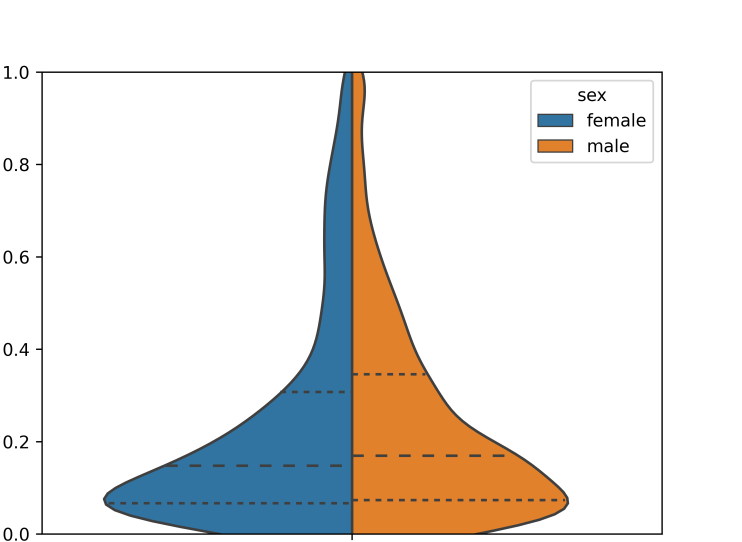} &
\distrimage{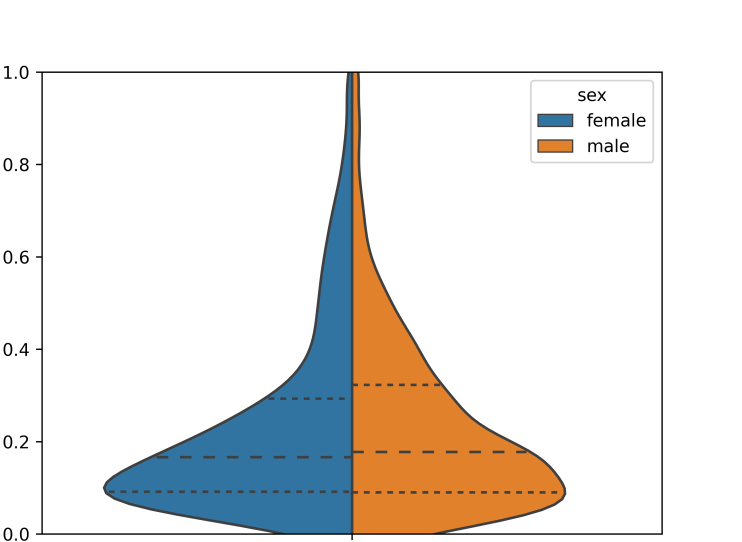} &
\distrimage{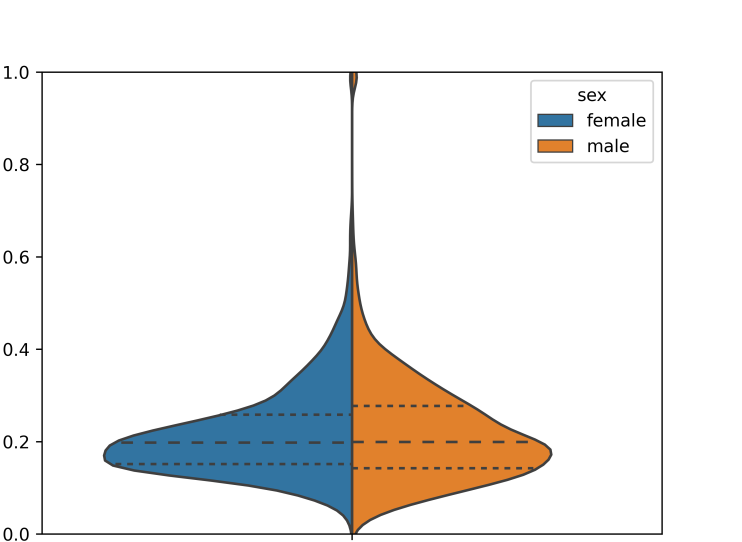} \\
$\epsilon = 0.01$ & 
\distrimage{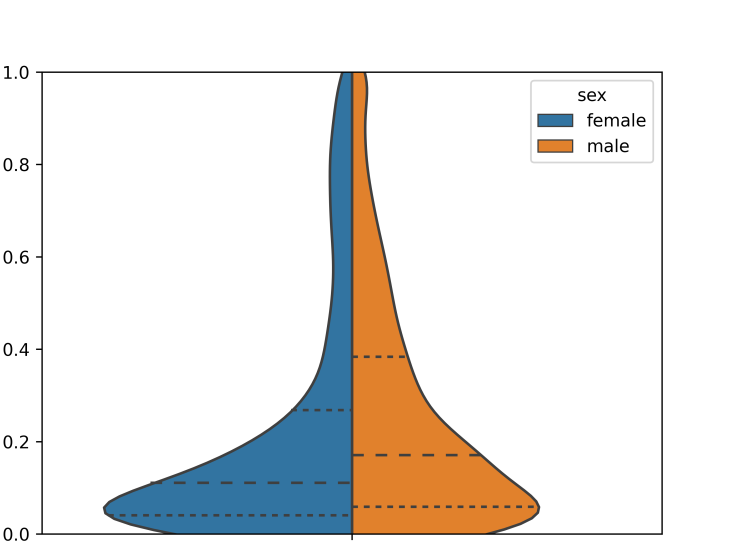} &
\distrimage{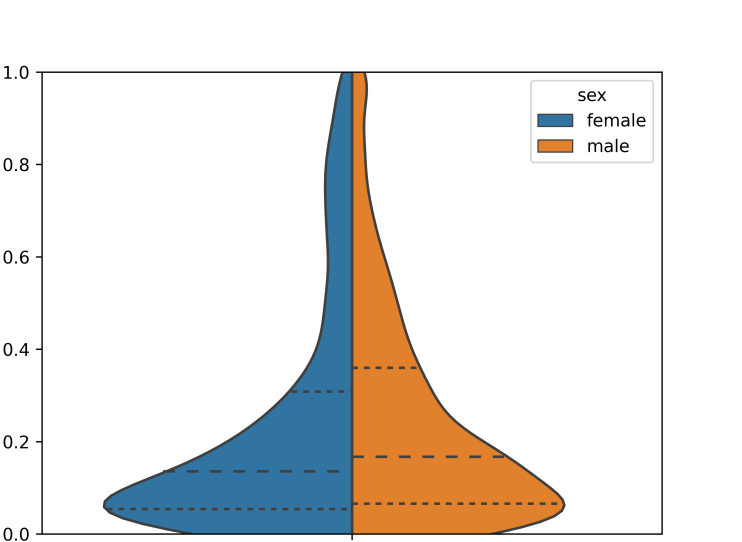} &
\distrimage{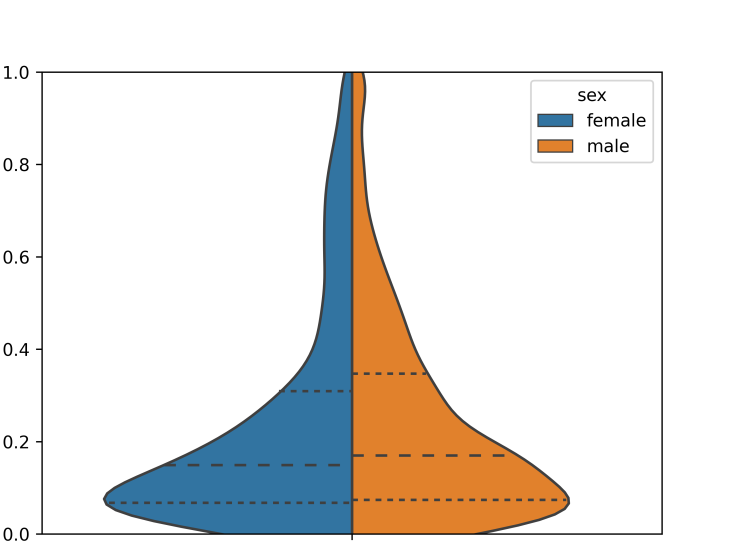} &
\distrimage{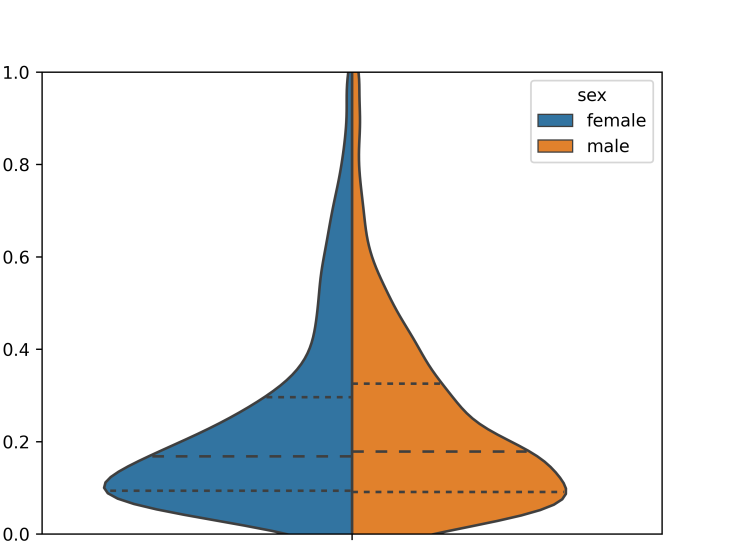} &
\distrimage{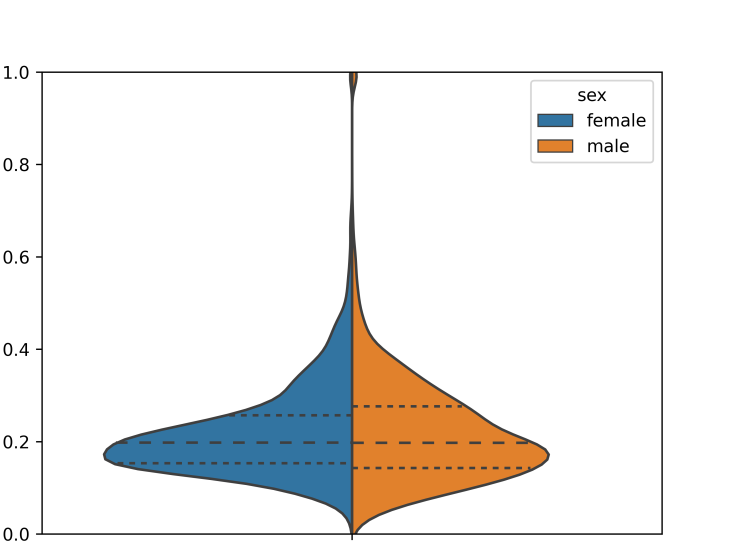} \\
$\epsilon = 0.1$ & 
\distrimage{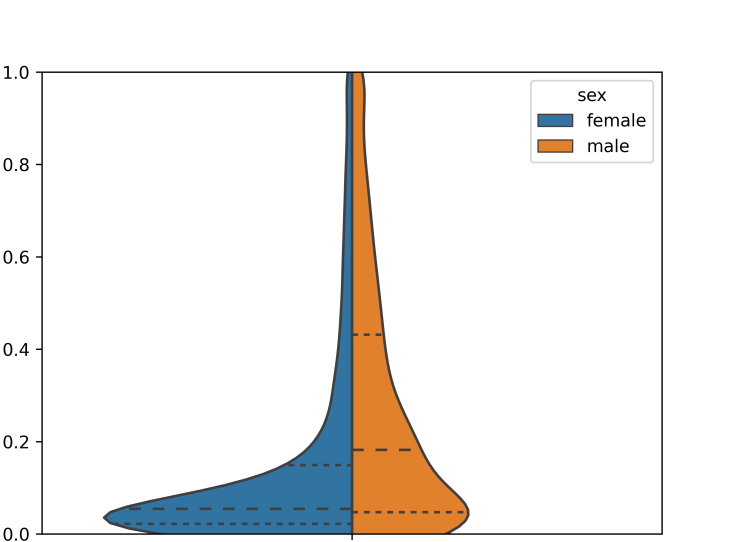} &
\distrimage{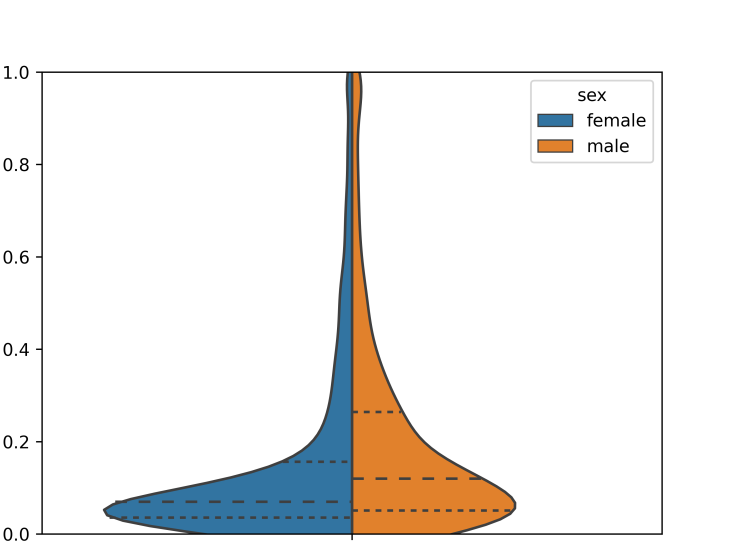} &
\distrimage{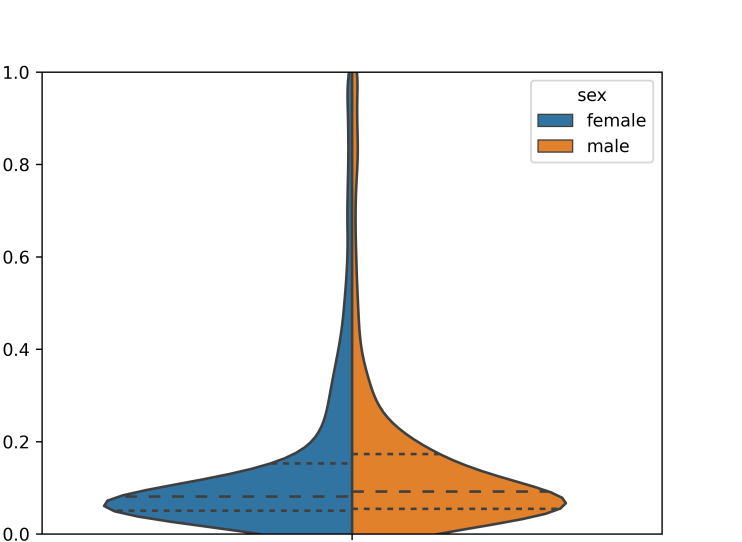} &
\distrimage{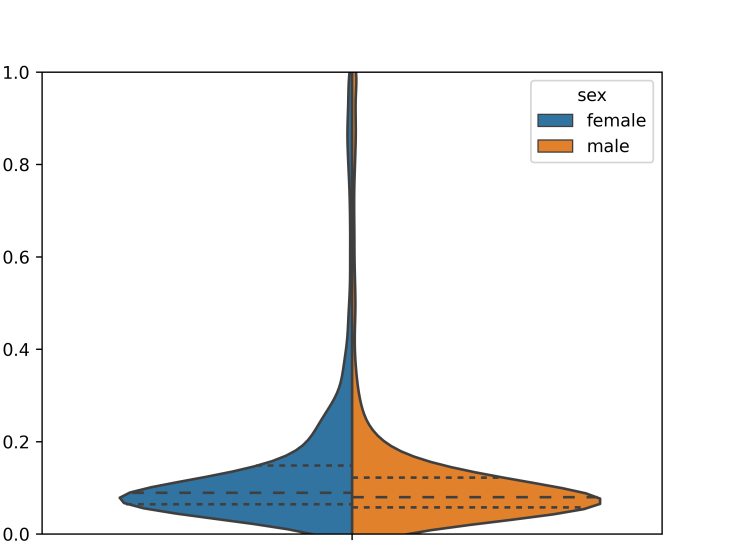} &
\distrimage{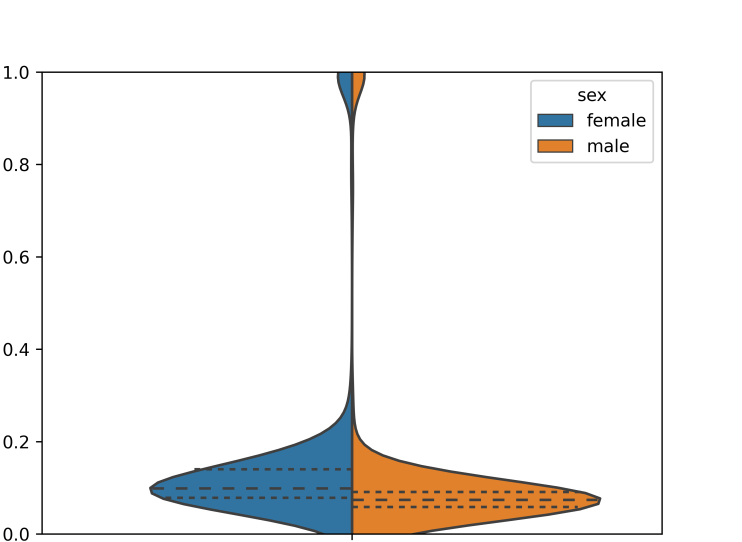} \\
$\epsilon = 1$ & 
\distrimage{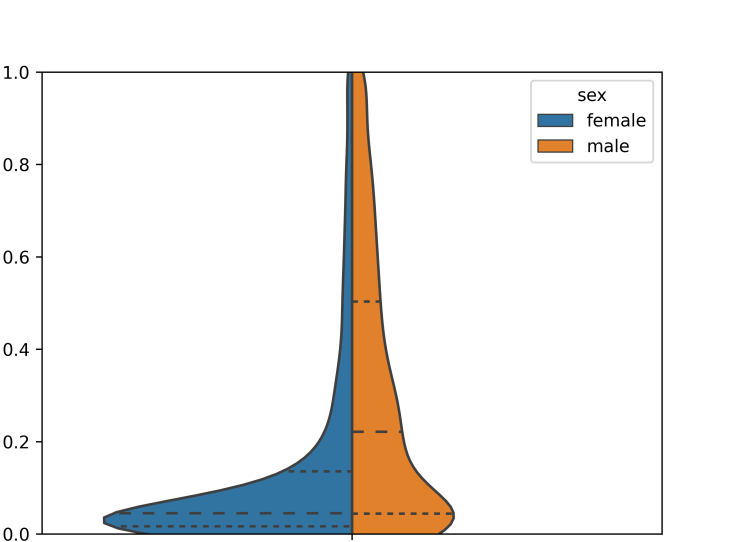} &
\distrimage{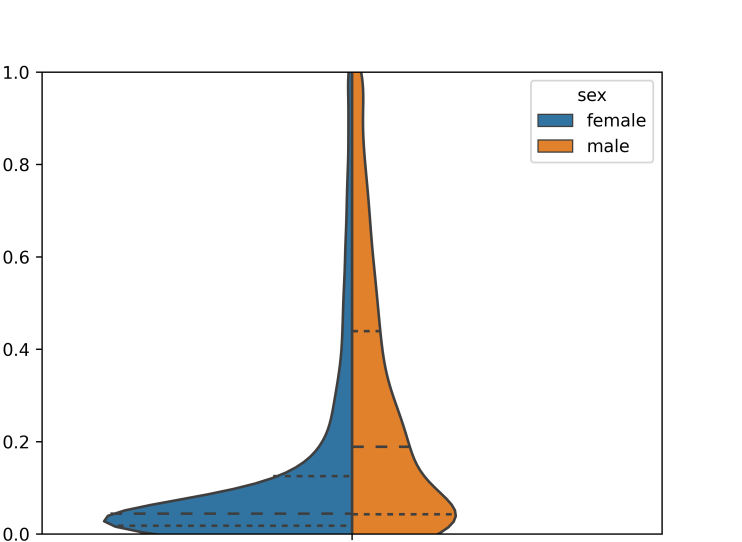} &
\distrimage{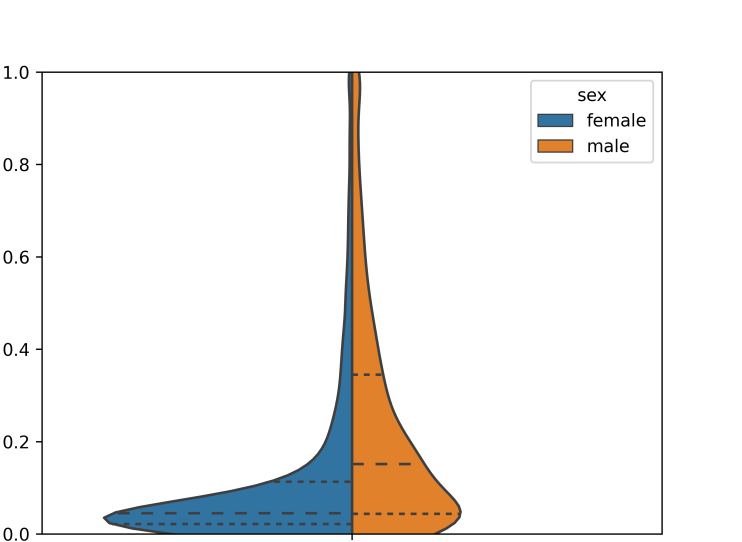} &
\distrimage{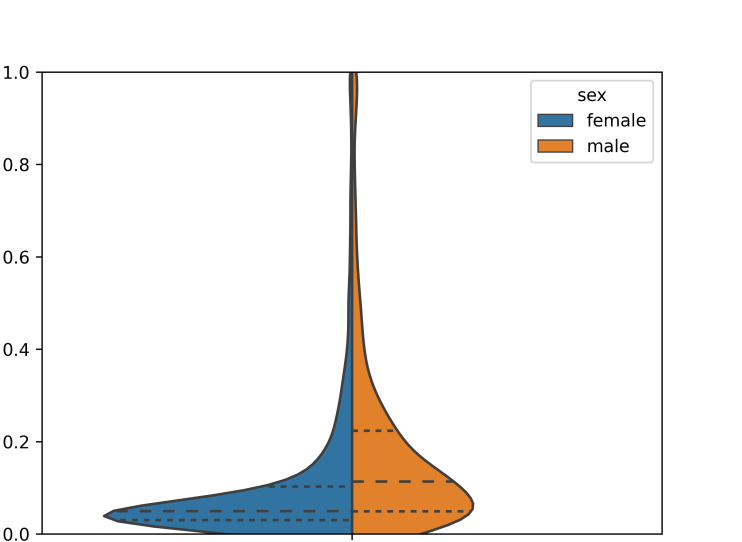} &
\distrimage{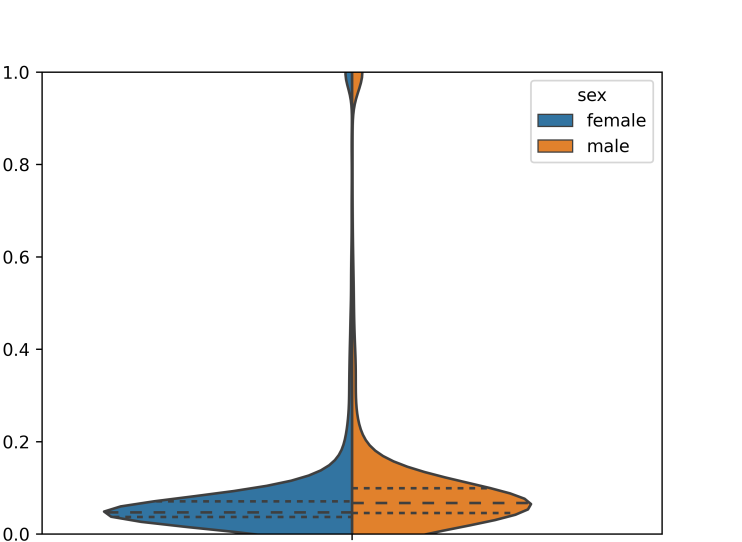} \\
$\epsilon = 10$ & 
\distrimage{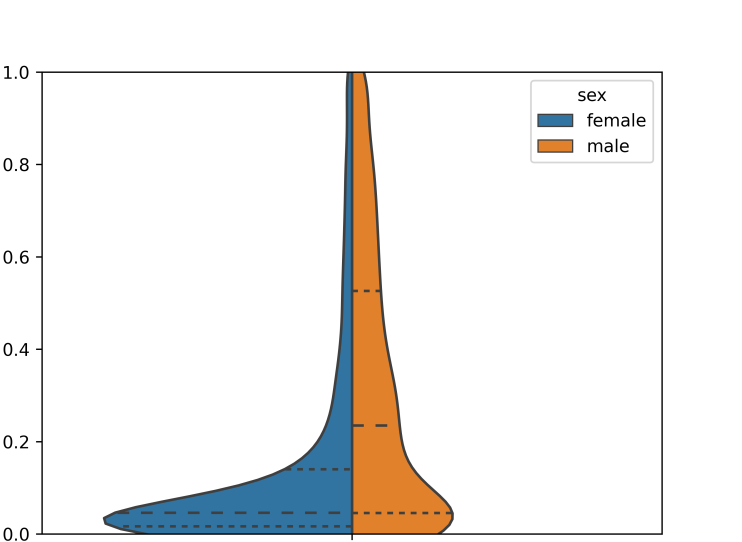} &
\distrimage{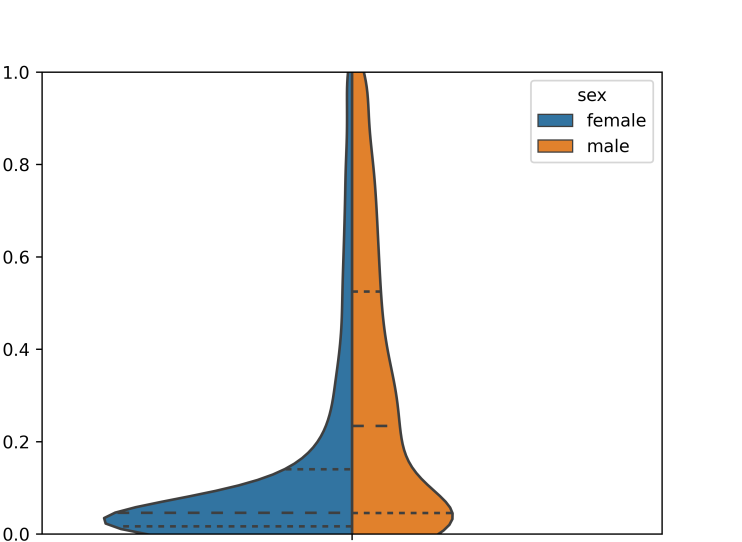} &
\distrimage{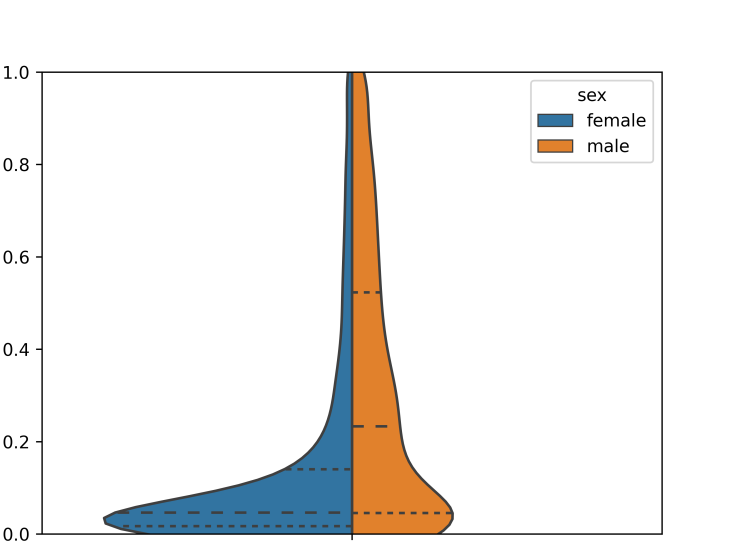} &
\distrimage{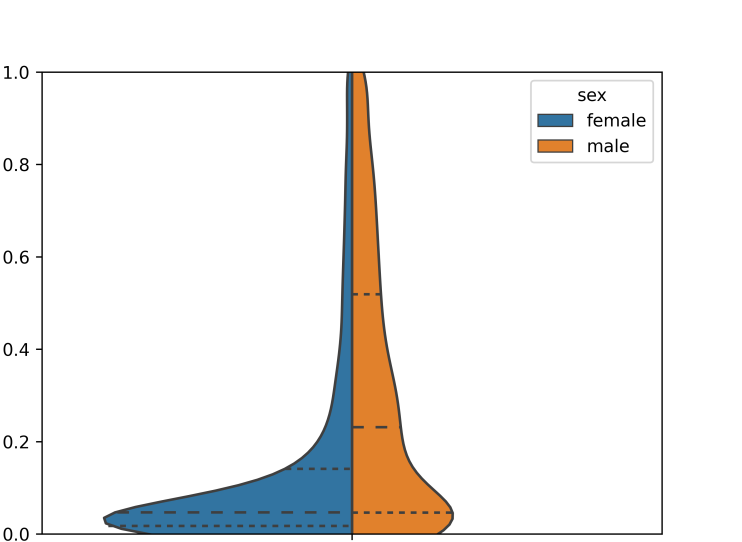} &
\distrimage{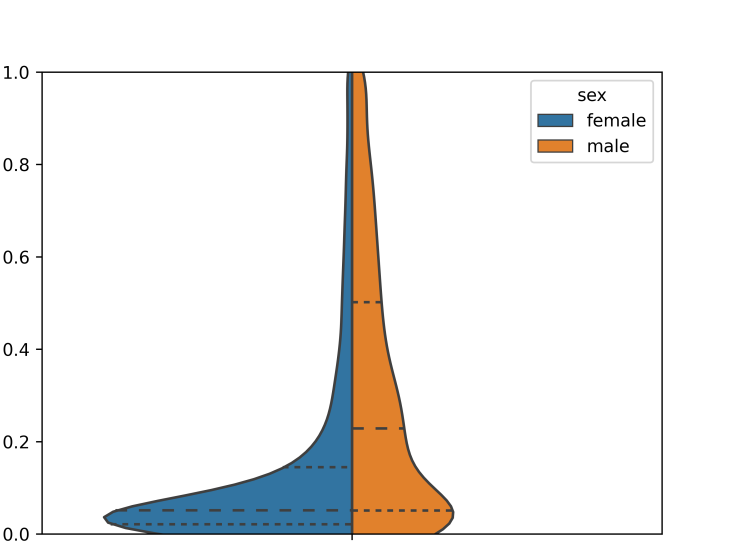} \\
\end{tabular}
\caption{For a model jointly trained with the $\tOTFe$ regularizer for different $\epsilon$ and $\alpha$ configurations, these are the violin plots of the probability scores for samples with \textsc{sex} attribute \textit{female} (blue) versus \textit{male} (orange). For each plot, the x-axis was normalized such that the maximum horizontal deviation from the center is constant. Dashed lines show the quartiles.}
\label{fig:scores}
\end{figure}

We aim to provide some empirical intuition for how the smoothing factor $\epsilon$ impacts the use of the $\tOTFe$ cost as a fairness regularizer. For this experiment, we used $\tOTFe$ for varying strengths of $\alpha$ and $\epsilon$ in order to minimize the violation of PDP with respect to only the \textsc{sex} attribute in the Adult dataset. All other settings are the same as those described in Sec.~\ref{sec:experiments}. The score distributions of these configurations are reported in Fig.~\ref{fig:scores}. 

It can be seen that for $\epsilon = 0.001$ and $\epsilon = 0.01$ the score distributions are made more similar for stronger $\alpha$ values. However, some properties, such as the more noticeable 'peak' for the samples in the female group compared to the male group, are maintained to some extent. 

For $\epsilon = 0.1$ and $\epsilon = 1$, the scores appear to be squished to very low or very high values, and they do not appear fair. We hypothesize that the relatively high value of the entropy term in the $\OTFe(h)$ and $\OTFRe(h)$ objectives overshadows the $\langle \bC, \bP \rangle$ term, which assigns a cost to how much score mass needs to be moved to make $h$ fair. Thus, the model may try to minimize $\tOTFe(h)$ mainly by minimizing the entropy term. This can indeed be accomplished by assigning very low or high element-wise probability scores.

For the even stronger smoothing with $\epsilon = 10$, we see that the fairness regularizer has no impact at all, because changes in $\alpha$ do not impact the score distributions that are eventually learned. For such a high $\epsilon$, both the $\OTFe(h)$ and $\OTFRe(h)$ solutions almost only strive to maximize entropy, causing these costs to cancel out. This causes $\tOTFe(h)$ to be close to zero even though $h$ itself may not yet be fair. 

We conclude that high $\epsilon$ values should be avoided, as the strong smoothing causes the unfairness signal in $\tOTFe(h)$ to be lost. It is then longer interesting as a fairness regularization term. 

\setlength{\tabcolsep}{1pt}

\subsection{Train Set Results for the Main Experiment}\label{sec:train_res}
\begin{figure}
	\centering
	\begin{subfigure}[b]{\textwidth}
		\centering
		\includegraphics[width=\textwidth]{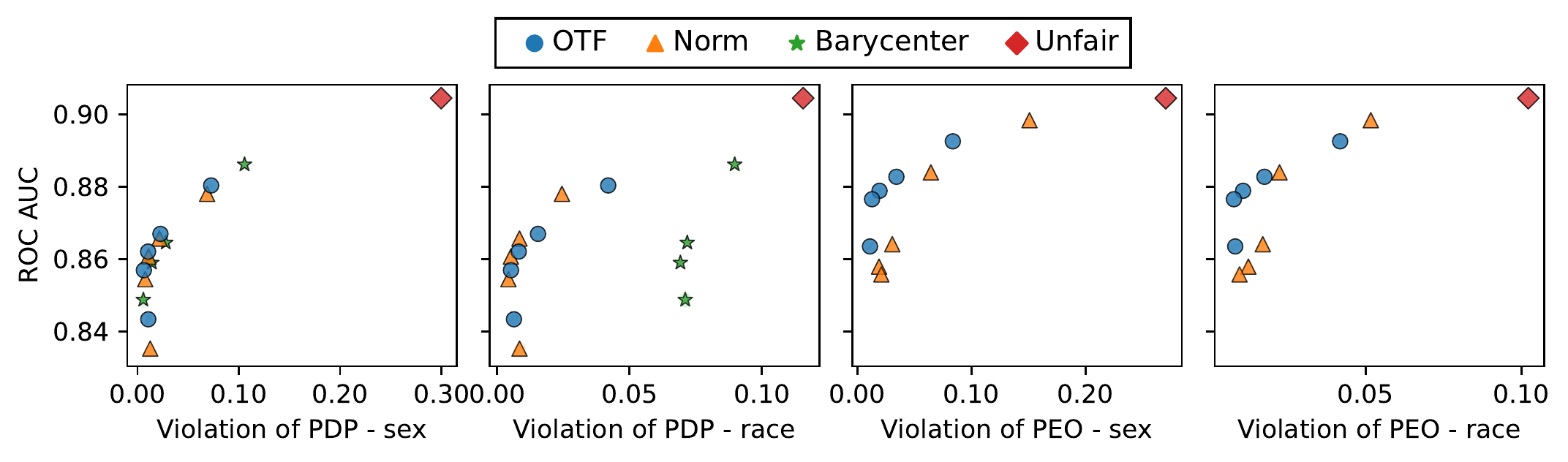}
		\caption{Adult dataset}
	\end{subfigure}
	\hfill
	\begin{subfigure}[b]{\textwidth}
		\centering 
		\includegraphics[width=\textwidth]{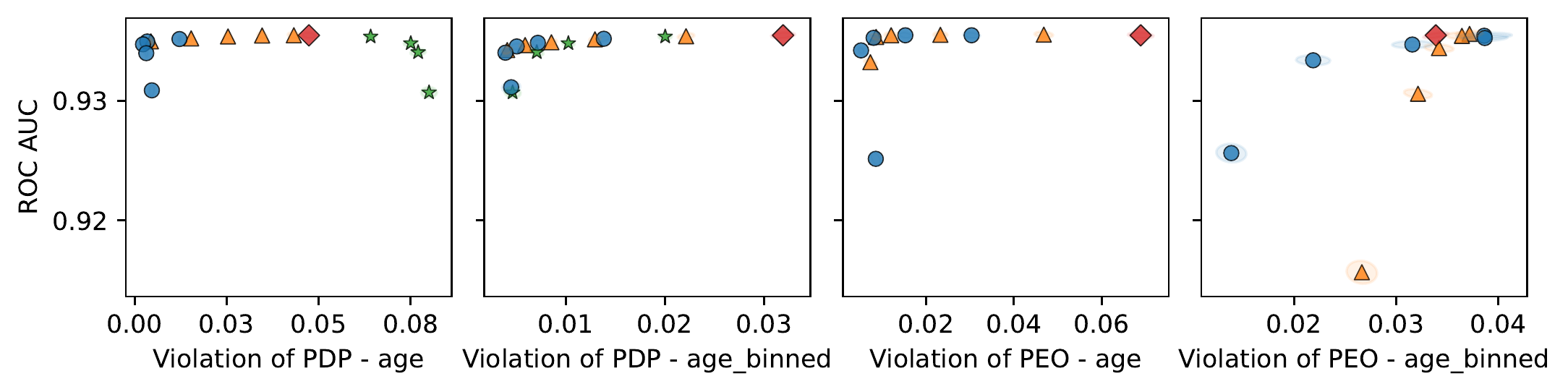}
		\caption{Bank dataset}
	\end{subfigure}
	\begin{subfigure}[b]{0.49\textwidth}
		\centering 
		\includegraphics[width=\textwidth]{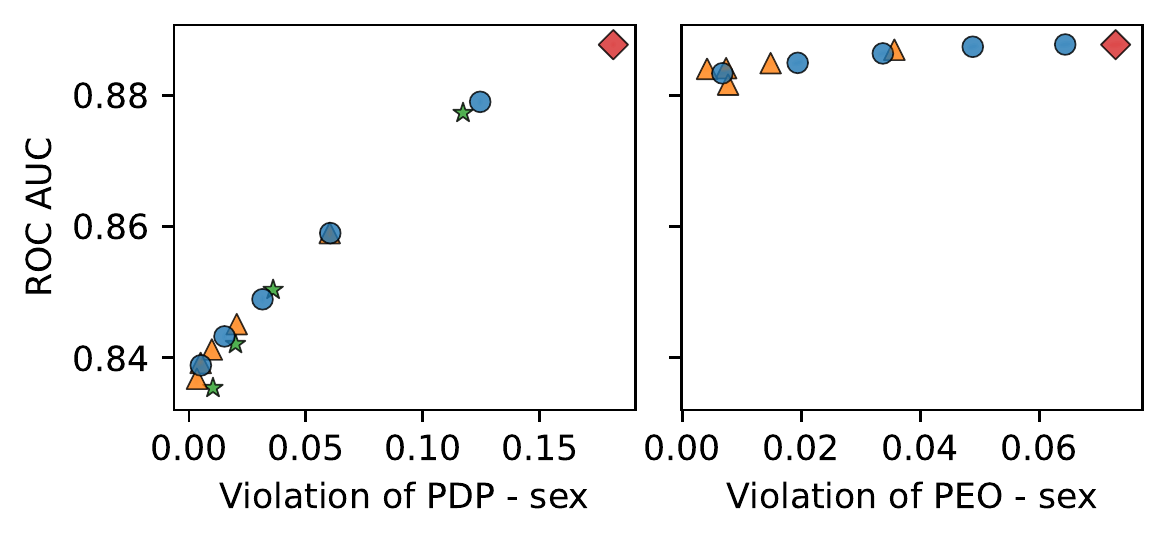}
		\caption{Dutch dataset}
	\end{subfigure}
	\begin{subfigure}[b]{0.503\textwidth}
		\centering 
		\includegraphics[width=\textwidth]{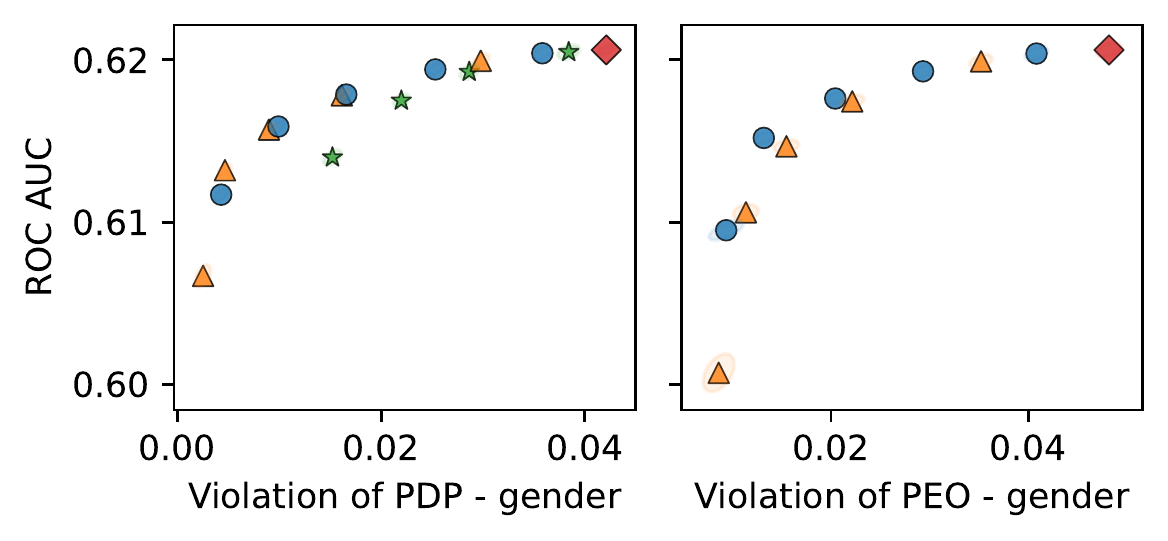}
		\caption{Diabetes dataset}
	\end{subfigure}
	\caption{Train set results for the methods that were trained to reduce the evaluated fairness measure (PDP or PEO). Violation of PDP (and PEO) is computed as the maximal absolute Pearson correlation between the probability scores (conditioned on the output labels) and each sensitive attribute.}
	\label{fig:exps_train}
\end{figure}

As discussed in Sec.~\ref{sec:eval}, we only report the test set results in the main paper. Results on the train set, which is far larger, are shown in Fig~\ref{fig:exps_train}. They follow the same mean trends but are less noisy due to the larger amount of samples. We thus draw the same conclusions as for the test set results discussed in Sec.~\ref{sec:results}. 

\section{Additional Clarification}
\subsection{Tying Measures to Score Functions}\label{sec:tying}
A score function $f: \mathcal{X} \to [0, 1]$ is involved in two domains: it operates on elements from the input space $\mathcal{X}$, but then produces a probability for value $1$ in the output space $\{0, 1\}$. In some prior work on OT for classifiers, \cite{frognerLearningWassersteinLoss2015, jiangWassersteinFairClassification2020}, the OT problem was posed using measures and a cost function over the \textit{output} space. Yet, our intention is to avoid transports between inputs that are highly dissimilar. We therefore tie classifiers to measures over the \textit{input} space $\mathcal{X}$ endowed with the Borel $\sigma$-algebra:
\begin{equation}\label{eq:input_space_measure}
	\theta_f(E) \triangleq \sum_{x \in \mathcal{D_X}} f(x)\delta_x(E)
\end{equation}
with $E \subseteq \mathcal{X}$, $\delta_e$ the Dirac measure (i.e. $\delta_e(E) = 1$ if $e \in E$, else $\delta_e(E) = 0$) and $\mathcal{D_X}$ all input features of samples in the dataset $\mathcal{D}$, gathered from the sample space $\mathcal{Z}$. Note that the input space measure $\theta_f$ is not normalized (i.e. $\theta_f(\mathcal{X}) \neq 1$), though this is not necessary to apply OT theory, which is most generally defined without any constraints on the total mass of the measures \cite{peyreComputationalOptimalTransport2019}.

In our formulation, we use Eq. (\ref{eq:input_space_measure}) to implicitly consider the score functions $h$ and $f$ as their corresponding input space measures $\theta_h$ and $\theta_f$ when used in the OT problem.

\subsection{Datasets}\label{sec:data_suppl}
All datasets are well-known in fairness literature \cite{lequySurveyDatasetsFairnessaware2022} and highly anonymized. 
The datasets are popularly used 
Both datasets are some of the most popular datasets hosted by the UCI repository \cite{duaUCIMachineLearning2017}. The data is highly anonymized.

For the Adult\footnote{Retrieved from \url{https://archive.ics.uci.edu/ml/datasets/adult}} dataset, gathered from the American Housing Survey. The task is to predict whether an individual earns more than \$50K/yr. We follow the default data preprocessing implemented by the \textit{AI Fairness 360}\footnote{\url{https://aif360.readthedocs.io}} framework and retain 45222 samples. The sensitive features are simplified to two binary sensitive attributes: \textsc{sex} (with values $\{\text{male}, \text{female}\}$) and \textsc{race} (with values $\{\text{white}, \text{non-white}\}$). 
%In the Adult\footnote{Retrieved from \url{https://archive.ics.uci.edu/ml/datasets/adult}} dataset, the task is to predict whether an individual earns more than \$50K/yr. We followed the default data preprocessing implemented by the \textit{AI Fairness 360}\footnote{\url{https://aif360.readthedocs.io}} framework, which retains 45222 samples. The sensitive features are simplified to two binary sensitive attributes: \textsc{sex} (with values $\{\text{male}, \text{female}\}$) and \textsc{race} (with values $\{\text{white}, \text{non-white}\}$). 

In the Bank\footnote{Retrieved from \url{https://archive.ics.uci.edu/ml/datasets/bank+marketing}} dataset \cite{moroDatadrivenApproachPredict2014}, the target is whether a client will subscribe to a product offered by a bank. For the 41188 data samples of individuals, the sensitive attribute is the age of the clients, which is traditionally converted to a categorical value by dividing the age into a limited number of bins. As sensitive attributes, we study both the original continuous \textsc{age} values and the quantized version based on the median age of 38, i.e. \textsc{age\_binned} (with values $\{<38, \geq38\}$).

The Dutch Census dataset\footnote{Retrieved from \url{https://github.com/tailequy/fairness_dataset/tree/main/Dutch_census}} \cite{vanderlaan2001} involves predicting whether the occupation of individuals is classified as `prestigious' or not. We followed the preprocessing outlined in \cite{lequySurveyDatasetsFairnessaware2022} and end up with 60420 samples. For this dataset, we only consider one sensitive attribute, the binary \textsc{sex} with values $\{\text{male}, \text{female}\}$. 

Finally, samples in the Diabetes dataset\footnote{Retrieved from https://archive.ics.uci.edu/ml/datasets/diabetes+130-us+hospitals+for+years+1999-2008} \cite{strackImpactHbA1cMeasurement2014}, represents features of patients for whom it should be predicted whether they will be readmitted within 30 days. Again following the preprocessing from \cite{lequySurveyDatasetsFairnessaware2022}, we end up with 45715 samples and use the listed \textsc{gender} of the patient as sensitive attribute with values $\{\text{male}, \text{female}\}$. 

\subsection{Hyperparameters}\label{sec:hyper}
For all methods, we used an unregularized, logistic regression model as the probabilistic classifier. We trained for 100 epochs with a learning rate of $10^{-3}$ and a batch size of 1000. The sensitive features were not included in the input $X$ to the model. For the OTF and Norm methods, we evaluated fairness regularization strengths $\alpha \in \{0.1, 0.3, 0.5, 0.7, 0.9\}$. 

Also for OTF, we chose $\epsilon = 10^{-3}$ from $\{10^{-4}, 10^{-3}, 10^{-2}, 10^{-1}\}$. We chose this value because it consistently resulted in an effective trade-off between AUC and fairness. For a discussion on the impact of setting $\epsilon$ too high, we refer to Sec.~\ref{sec:eps_disc}.

For the Barycenter method, we use the Wasserstein-1 distance for the penalized logistic regression mentioned in Eq. (3) of  \cite{jiangWassersteinFairClassification2020}. We set $\beta = 1000$ and used fairness strengths $\alpha \in \{0.01, 0.05, 0.1, 0.2\}$.

\end{document}